\documentclass[
]{ceurart}

\usepackage{listings}

\usepackage{microtype}
\usepackage{graphicx}
\usepackage{subcaption}
\usepackage{booktabs}
\usepackage{url}
\usepackage{dirtytalk}
\usepackage{tcolorbox}
\usepackage{listings}
\usepackage{xcolor}
\definecolor{myxcolor}{RGB}{220, 53, 69}
\usepackage{tikz}
\usepackage{pgfplots}
\usepackage{makecell}

\usepackage{algpseudocode}
\usepackage{multirow}
\usepackage{hyperref}
\usepackage{pifont}
\newcommand{\cmark}{\ding{51}}
\newcommand{\xmark}{\ding{55}}
\usepackage{placeins}
\usepackage[romanian,spanish, russian, english]
{babel}

\usepackage{algorithm}
\usepackage{algpseudocode}
\usepackage{multirow}
\usepackage{hyperref}
\usepackage{changepage}

\begin{document}

\usetikzlibrary{math}

\copyrightyear{2026}
\copyrightclause{Copyright for this paper by its authors.
  Use permitted under Creative Commons License Attribution 4.0
  International (CC BY 4.0).}

\conference{}

\title{ML$^2$B: Benchmarking LLMs on Cross-Lingual ML Pipeline Generation}


\author[1, 2]{Ekaterina Trofimova}[%
orcid=0000-0001-5436-8511,
email=ekaterina.trofimova@constructor.org,
]
\cormark[1]
\address[1]{Constructor Labs, Campus Ring 1, 28759 Vegesack, Bremen, Germany}

\author[2]{Zosya Shamina}[%
orcid=0009-0000-2870-4403,
email=zosyasham@gmail.com,
]
\address[2]{HSE University, Moscow, Russia}

\author[2]{Maria Selifanova}[%
orcid=0009-0009-7065-0512,
email=mariaselifanova1@gmail.com,
]

\author[2]{Artem Zaitsev}[%
orcid=0009-0002-0286-6484,
email=arvlzaitsev@gmail.com,
]

\author[2]{Remi Savchuk}[%
orcid=0009-0007-9083-0292,
email=enaix@protonmail.com,
]

\author[2]{Maxim Minets}[%
orcid=0009-0004-5895-1391,
email=m.maxim.v.01@gmail.com,
]

\author[2]{Daria Ozerova}[%
orcid=0009-0002-5423-6409,
email=dozerova01@gmail.com,
]

\author[2]{Emil Sataev}[%
orcid=0009-0000-2891-0792,
email=emils99mail@gmail.com,
]

\author[2]{Denis Zuenko}[%
orcid=0000-0002-6148-5107,
email=neron.v2@gmail.com,
]

\author[1, 3]{Andrey Ustyuzhanin}[%
email=austyuzhan@constructor.university,
]
\address[3]{National University of Singapore, 21 Lower Kent Ridge Rd, University Hall, 119077, Singapore}

\cortext[1]{Corresponding author.}





\begin{abstract}
We introduce ML$^2$B, the first benchmark for evaluating 
cross-lingual task comprehension in end-to-end ML pipeline 
generation by large language models. Despite growing global AI adoption, 
no systematic evaluation exists for ML pipeline generation beyond English 
task descriptions. ML$^2$B addresses this gap with 35 Kaggle competitions 
spanning tabular, text, and image domains, translated into 14 languages 
by native-speaker researchers with ML expertise, yielding 490 
task-language pairs. To ensure evaluation integrity, the benchmark 
incorporates 10 private competitions without publicly available solutions 
and employs network-isolated evaluation infrastructure restricting runtime 
access to essential ML resources. We provide standardized evaluation 
protocols, an AutoGluon algorithmic baseline, and comprehensive failure 
mode analysis. Experiments with frontier models (GPT-4.1-mini, 
GPT-OSS-120b, Gemini-2.5-Flash) reveal that cross-lingual performance 
degradation is highly task-dependent rather than following traditional 
resource-availability hierarchies, with gaps ranging from language 
advantages to severe degradation depending on competition 
characteristics. These findings challenge conventional assumptions about 
multilingual model capabilities and underscore the necessity of systematic 
cross-lingual evaluation for ML pipeline generation. We open-source the benchmark, baselines, and evaluation infrastructure at \url{https://github.com/enaix/ml2b}.
\end{abstract}

\begin{keywords}
Multilingual machine learning  \sep
large language models  \sep
cross-lingual representation learning  \sep
code generation  \sep
machine learning work-flows  \sep
benchmark dataset
\end{keywords}

\maketitle

\section{Introduction}

Machine learning (ML) has become a fundamental component across various 
domains, motivating the development of AutoML frameworks to automate 
pipeline selection~\citep{10.1613/jair.1.11854} and LLM-based code 
generation benchmarks such as MLE-Bench~\citep{chan2025mlebench}, 
DA-Code~\citep{huang-etal-2024-da}, and Weco-Kaggle~\citep{jiang2025aide}. 

\begin{table}[h!]
\caption{Comparison of ML$^2$B with related benchmarks: leakage
detection, tasks without public solutions (reducing training-data
contamination), assessment of specific ML capabilities, and grading
in a network-isolated container.}
\centering
\small
\resizebox{\linewidth}{!}{
\begin{tabular}{lccccc}

\hline
\textbf{Benchmark} & \textbf{Multilingual} & \textbf{Leakage Prevention} & \textbf{Private Competitions} &
\textbf{ML Capabilities} &
\textbf{Isolated Grading} \\
\hline
ML$^2$B (ours)          & \textcolor{teal}{\cmark} & \textcolor{teal}{\cmark} & \textcolor{teal}{\cmark} & \textcolor{myxcolor}{\xmark} & \textcolor{teal}{\cmark} \\
MLE-Bench            & \textcolor{myxcolor}{\xmark} & \textcolor{myxcolor}{\xmark} & \textcolor{myxcolor}{\xmark} & \textcolor{myxcolor}{\xmark} & \textcolor{myxcolor}{\xmark} \\
ML-Dev-Bench         & \textcolor{myxcolor}{\xmark} & \textcolor{myxcolor}{\xmark} & \textcolor{teal}{\cmark} & \textcolor{teal}{\cmark} & \textcolor{myxcolor}{\xmark} \\
MLAgentBench         & \textcolor{myxcolor}{\xmark} & \textcolor{myxcolor}{\xmark} & \textcolor{myxcolor}{\xmark} & \textcolor{teal}{\cmark} & \textcolor{myxcolor}{\xmark} \\
DSCodeBench          & \textcolor{myxcolor}{\xmark} & \textcolor{myxcolor}{\xmark} & \textcolor{myxcolor}{\xmark} & \textcolor{teal}{\cmark} & \textcolor{myxcolor}{\xmark} \\
\hline
\end{tabular}
}

\label{tab:mlben_comparison}
\end{table}

All existing benchmarks are limited to English. LLMs exhibit systematic 
performance degradation in other languages~\citep{li2025language, qin2025survey}, 
with substantial drops for low-resource 
languages~\citep{Xuan2025MMLU-ProX, raihan-etal-2025-mhumaneval}. 
Meanwhile, ML research is global: models must interpret problem descriptions 
in diverse languages while producing executable code. Current benchmarks 
cannot measure this cross-lingual robustness.

Another challenge arises from \emph{benchmark data leakage}, when benchmark data is also present in the LLM training data~\citep{matton2024leakage}. This issue is particularly important, as the model may overperform in particular benchmark tasks. In the worst-case scenario, this may lead to the affected benchmark competitions being inconclusive~\citep{zhou2025lessleakbench}. A similar form of data leakage happens when unintended information about the test set appears in training data. Leakage can artificially inflate performance and produce unreliable results~\citep{Apicella_2025, yang2022dataleak, Sasse_2025}. It is also pervasive in real-world ML code~\citep{leakage_in_sci}.

To address these shortcomings, we introduce ML$^2$B, the first benchmark 
for evaluating LLMs on generating complete ML pipelines from multilingual 
natural language descriptions. Our key contributions are:

\begin{enumerate}
  \item \textbf{Multilingual benchmark.} 35 Kaggle competitions translated 
  into 14 natural languages, yielding 490 evaluation instances.
  \item \textbf{Private competitions.} 10 competitions without publicly 
  available solutions, reducing the likelihood of training data contamination.
  \item \textbf{Robust code evaluation.} Network-isolated grading, modular 
  submission format, and static leakage detection.
\end{enumerate}

Beyond the language dimension, ML$^2$B differs from MLE-Bench and
DA-Code methodologically: agent code is executed directly via AST-based
recompilation rather than scored as a submission file, the modular
format structurally separates training from prediction data, and
grading runs in a network-isolated container
(Table~\ref{tab:mlben_comparison}).


\section{Related work}

\subsection{ML Code Generation and AutoML}

Recent benchmarks evaluate LLM capabilities for ML code generation across different scopes. DSCodeBench \citep{ouyang2025dscodebench} and DS-1000 \citep{ds100} assess snippet-level code from GitHub and StackOverflow. Full-pipeline benchmarks include DA-Code \citep{huang-etal-2024-da}, Weco-Kaggle \citep{jiang2025aide}, and MLE-bench \citep{chan2025mlebench}, which leverage Kaggle competitions. MLE-bench evaluates agents on 75 tasks; ML-Dev-Bench \citep{padigela2025mldevbenchcomparativeanalysisai} provides 30 tasks for debugging and API integration; MLAgentBench \citep{huang2024mlagentbenchevaluatinglanguageagents} offers 13 research-oriented scenarios. All are English-only.

Contemporary LLM-based agents for ML engineering employ iterative refinement strategies. AIDE \citep{jiang2025aide} uses solution space tree search, outperforming traditional AutoML frameworks like LightAutoML \citep{vakhrushev2022lightautomlautomlsolutionlarge} and OpenHands \citep{wang2025openhandsopenplatformai} on Kaggle tasks. Recent multi-agent frameworks like ML-Master~\citep{liu2025mlmasteraiforaiintegrationexploration} and R\&D-Agent~\citep{yang2025rdagentllmagentframeworkautonomous} achieve superior performance but require substantially larger models (GPT-4.1, GPT-5). 

\subsection{Multilingual Code Generation}

Multilingual datasets for code generation remain scarce. MCoNaLa~\citep{wang2022mconala} consists of intents for code generation, which are further rewritten by human annotators, and code snippets in Python. RoCode~\citep{cosma-etal-2024-rocode} offers Romanian programming problems with Python/C++ solutions. MBPP-Translated~\citep{li2024bridging} extends MBPP to five languages using machine translation. mHumanEval~\citep{raihan-etal-2025-mhumaneval} supports 204 languages, with expert translation for 15, across 25 programming languages. However, none target end-to-end ML pipelines.

Several studies show LLM performance depends strongly on prompt language. \citet{bang-etal-2023-multitask, ahuja2023mega, muennighoff-etal-2023-crosslingual}, and \citet{raihan-etal-2025-mhumaneval} report substantial drops for low-resource languages. \citet{moumoula2025evaluating} analyze 13 programming and 23 natural languages, showing that non-Latin scripts further degrade performance. While targeted multilingual fine-tuning can partially mitigate these issues \citep{huo2025enhancing, shaham2024multilingual}, current ML code-generation benchmarks provide no means to measure such cross-lingual robustness.

\section{The ML$^2$B benchmark}

ML$^2$B provides structured, succinct metadata and task descriptions 
designed for efficient LLM processing, unlike MLE-Bench~\citep{chan2025mlebench}, which uses verbose descriptions sourced directly from Kaggle. Structured prompts achieve better performance on 5 of 10 shared competitions and perfect stability (3/3 pass rate) versus obfuscated format failures on 5 competitions, confirming that clear task 
descriptions aid comprehension without enabling memorization-based 
shortcuts (Appendix~\ref{app:prompt_comparison}). The benchmark 
comprises 35 competitions spanning 14 languages (490 evaluation 
instances), with 10 private competitions to mitigate data contamination 
(Appendix~\ref{app:competition_list}).

\subsection{Benchmark task selection}
\label{sec:bench_task_selection}

The ML$^2$B benchmark builds on the Code4ML 2.0 dataset~\citep{ekaterina_trofimova_2024_12700065}, which provides standardized task descriptions for Kaggle competitions. We first identify all competitions meeting our inclusion criteria and released after 2020. From these, we manually validate and select 22 tasks with publicly available Kaggle code. Selected competitions satisfy the following conditions: (1) the competition is closed, ensuring fixed data and leaderboards; (2) the dataset is publicly downloadable; (3) the evaluation metric is documented; (4) the evaluation is reproducible with complete metadata; and (5) submissions follow a tabular prediction format.

As Code4ML 2.0 covers competitions up to 2024, we additionally include three public competitions released in 2025, resulting in 25 public tasks. To assess generalization to unseen problems, we further add 10 private competitions without publicly available code, i.e., finished, closed Kaggle competitions with no public solution code, no forum discussions, and limited web indexability (hosted by university courses or small organizations). These retain public leaderboards, enabling consistent baseline evaluation (Section~\ref{sec:evaluation}), but lack solution code and forums, reducing the likelihood of inclusion in LLM training data. Task descriptions are manually summarized from the corresponding Kaggle pages. 

All datasets and task descriptions are sourced from Kaggle competition pages and released 
under Kaggle’s standard competition license or permissive Creative Commons licenses 
(CC BY 4.0 or CC BY-NC 4.0), as specified on each competition page. These licenses allow 
redistribution with attribution, with CC BY-NC 4.0 restricting commercial use.

Unlike Code4ML 2.0, ML$^2$B provides a manually curated data card for each task in addition to the data source link. Data cards and task descriptions are manually reviewed to ensure clarity and to prevent information leakage that could confer an unfair advantage. In particular, details related to Kaggle submission formats and test files are removed, as test data lack labels and are irrelevant to our executable-code generation setting. Each task also includes its evaluation metric and metric type, following the taxonomy of~\citep{code4ml}.

\subsection{Benchmark Characteristics}
Table~\ref{tab:bench_stats} summarizes ML$^2$B coverage. Competitions span 9 domains (finance, healthcare and medical, manufacturing and quality, environmental studies, media studies, social studies, urban studies, marketing, synthetic data), 
ensuring broad coverage (Appendix~\ref{app:domain}). Evaluation metrics include 5 higher-is-better (AUC, F1, MAP, R², accuracy) and 3 lower-is-better (RMSE, MAE, log-loss) metrics. Overall, both the full and chosen sets share relatively similar characteristic distributions, which supports using the subset to present our results sustainably.

\begin{table}[t!]
\caption{Competition characteristics on full and sub-set also grouped by public vs. private partition. The private split is identical for the full set and the chosen one.}
\centering
\small
\begin{tabular}{llccc}
\hline
\textbf{Partition} 
& \textbf{Set} 
& \makecell{\textbf{Size}}
& \makecell{\textbf{Data type} (tab / text / img)}
& \makecell{\textbf{Metric direction} (higher / lower)} \\
\hline
Public  
  & Full set   & 25 & 20 / 2 / 3 & 14 / 11 \\
  & Chosen set & 5  & 3 / 1 / 1  & 3 / 2 \\
\hline
Private 
  & Full/chosen set & 10 & 8 / 1 / 1 & 5 / 5 \\
\hline
\end{tabular}

\label{tab:bench_stats}
\end{table}

\subsection{Metadata and Structure} 

\begin{figure*}[ht!]
    \begin{center}
    \includegraphics[width=\linewidth, trim=0 50 0 0, clip]{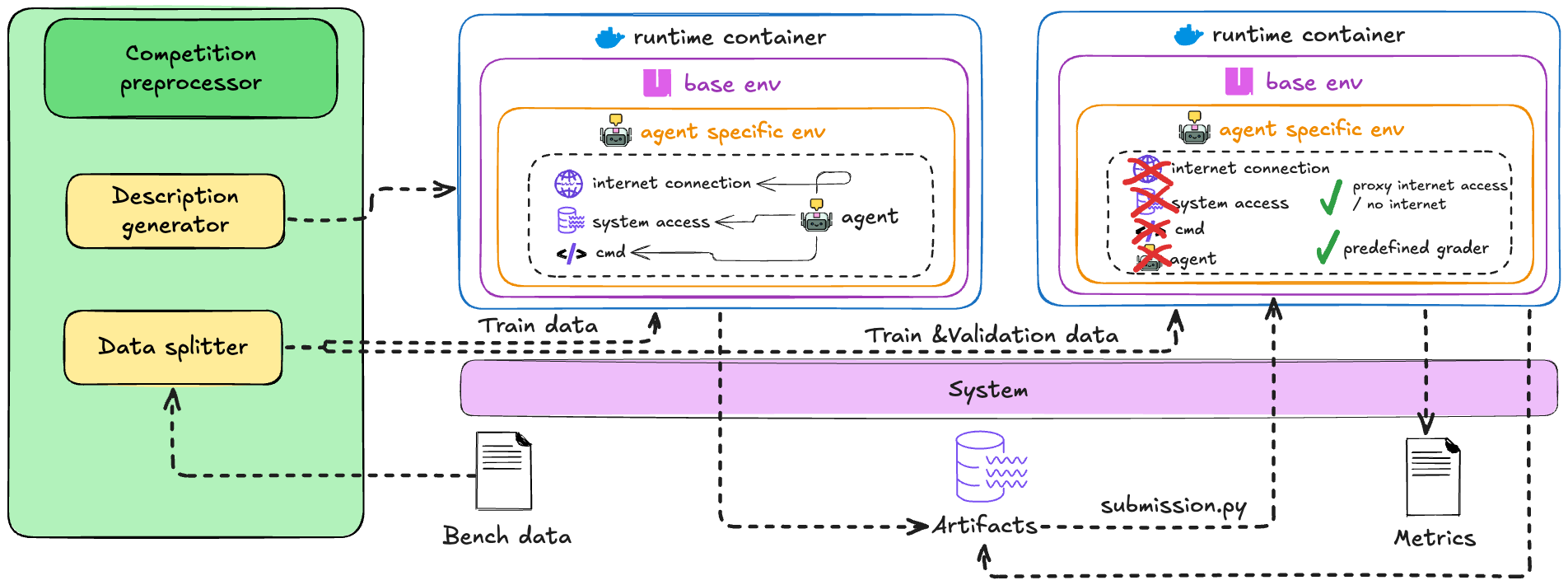}
    \end{center}
    \caption{Structure of the ML$^2$B benchmark. Competition preprocessor prepares task descriptions and data, runtime container generates the solution while evaluation container executes the solution code}
    \label{fig:bench_struct}
\end{figure*}

The benchmark consists of 3 main components (Figure \ref{fig:bench_struct}), which include the main competition preprocessor, Docker agent runtime, and the submission code grader. The competition preprocessor is responsible for task description generation (Appendix~\ref{app:prompt_example_competition}) and competition data preparation, the agent runtime manages the AutoML agent, and the code grader evaluates a metric for the submission code in an isolated environment.

Both the agent and code grader are executed inside the Docker environment. The agent Docker image is built from the common runtime image, and both the agent and grader containers are built from the same agent image. This ensures that both the agent and the grader utilize the same Python environment, and at the same time, grading is performed in an isolated environment without internet access. This prevents the potentially sensitive evaluation data from leaking in the event of a misconfigured or malicious script being submitted.

Agent runtime permits full internet access during code generation. 
The evaluation grader operates behind a forward proxy restricting 
access to essential ML infrastructure (HuggingFace, PyTorch, S3), 
blocking Kaggle and GitHub to prevent solution retrieval at runtime.

Instead of the Kaggle-style submission format, which consists of a single submission file, the code grader reproduces the results by executing the submission code directly. Furthermore, the code submitted by the agent must provide specific functions, which are then individually evaluated. Such an approach ensures that the submitted code is valid and can be reproduced in the controlled environment. In order to successfully load the submission code, the submission must not have top-level executable code. In order to achieve this, the grader must analyze and recompile the code by performing an Abstract Syntax Tree (AST) transformation. Then, the recompiled submission code is executed and the data is loaded into memory by the competition \texttt{DataLoader} class. Finally, the resulting submission data is evaluated using the corresponding competition grader function.

\begin{figure*}[ht!]
\begin{subfigure}{0.5\textwidth}
\centering
\includegraphics[height=0.25\textheight]{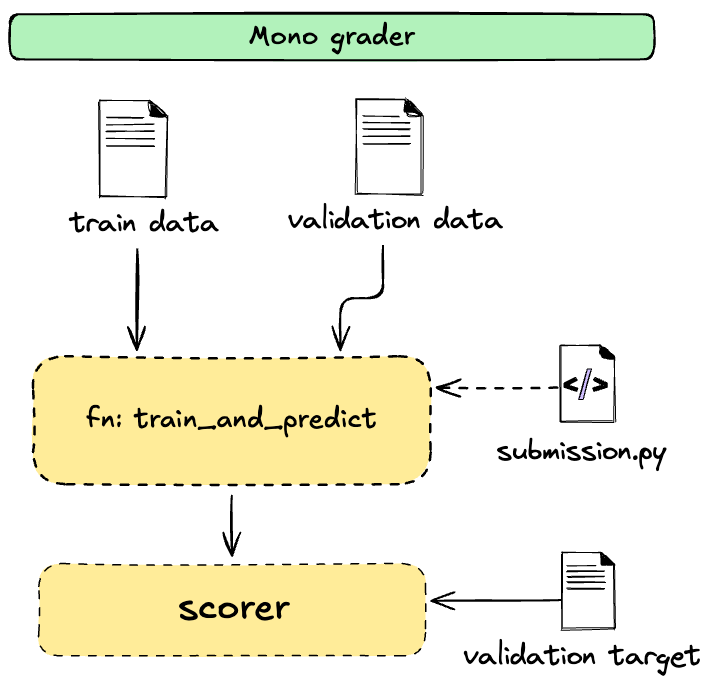}
\caption{Mono grader is the default code submission format, where a single-function solution returns the prediction}
\label{fig:sub_mono}
\end{subfigure}
\begin{subfigure}{0.45\textwidth}
\centering
\includegraphics[height=0.25\textheight]{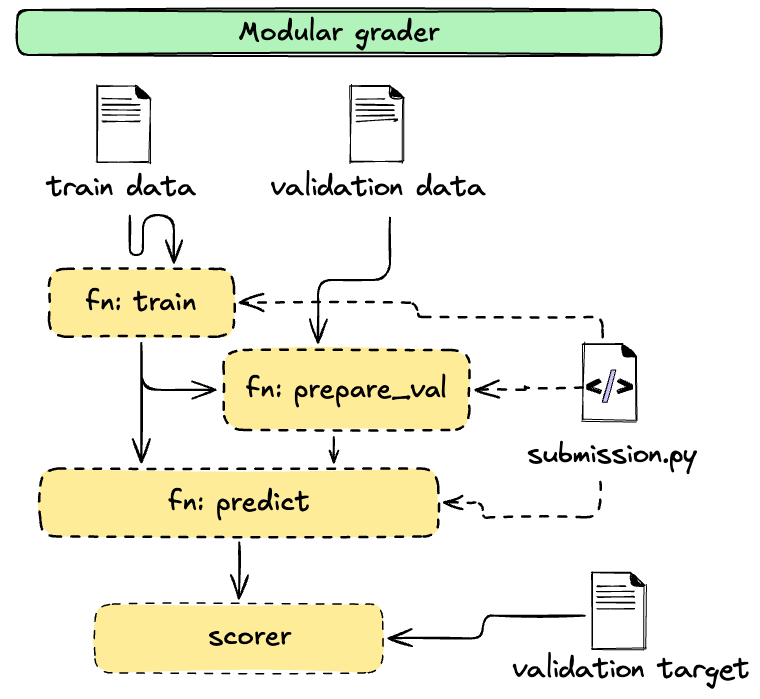}
\caption{Modular grader is the introduced format which isolates the train data from the validation data}
\label{fig:sub_modular}
\end{subfigure}
\caption{Code flow diagram of agent solution submission formats}
\label{fig:grader_struct}
\end{figure*}

\begin{figure*}[ht!]
    \begin{center}
    \includegraphics[width=0.75\linewidth]{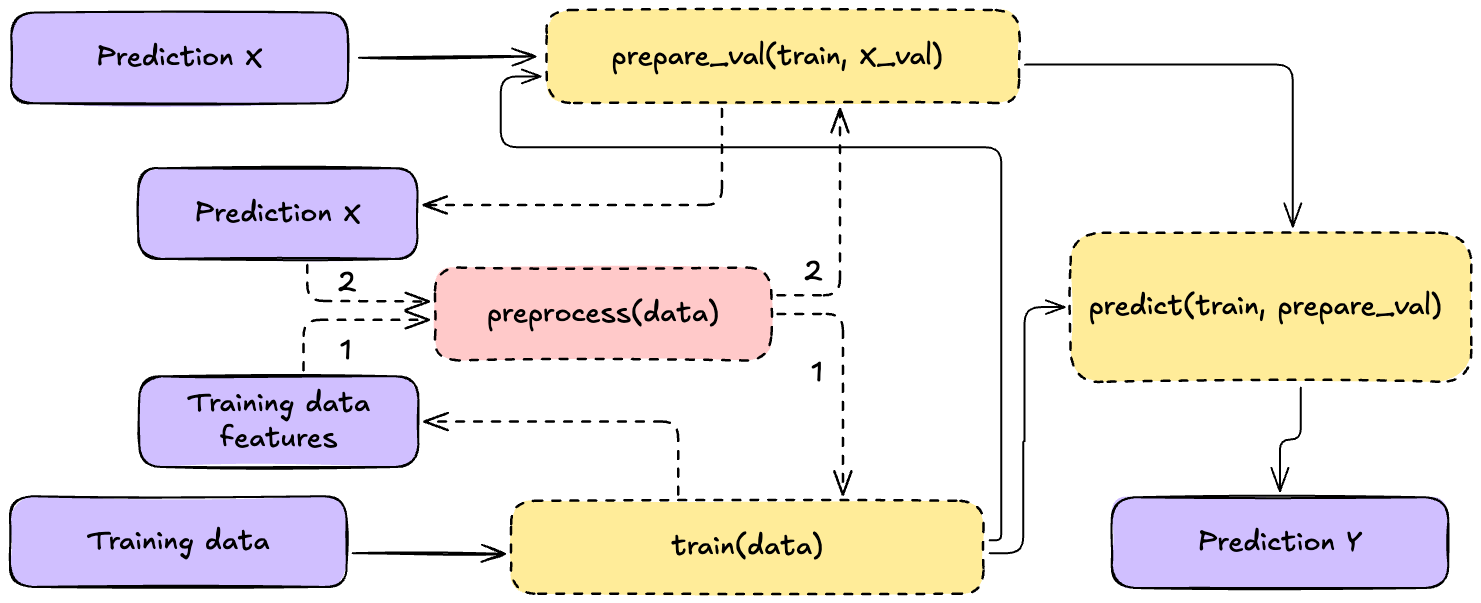}
    \end{center}
    \caption{Code flow diagram of a false-positive data leakage, where the preprocessing function is called sequentially}
    \label{fig:leak_data}
\end{figure*}

The benchmark supports two submission formats: single-function submission format \texttt{MONO\_PREDICT} (Figure~\ref{fig:sub_mono}) and modular submission format \texttt{MODULAR\_PREDICT} (Figure~\ref{fig:sub_modular}). \texttt{MODULAR\_PREDICT} consists of three functions \texttt{train}, \texttt{prepare\_val}, and \texttt{predict}, which sequentially train the model, prepare the prediction data, and predict the result. The purpose of such a prediction format is to reduce the chance of preprocessing leakage. Preprocessing leakage is the type of data leakage when both training and test data are processed together~\citet{yang2022dataleak, Apicella_2025}. The most common example is the data normalization being trained on both the training and test features. \texttt{MODULAR\_PREDICT} format restricts the code flow in a way that the prediction data is introduced only in the second stage of the pipeline, which makes the occurrence of preprocessing leakage less likely. In order to assess the presence of data leakage in submission code, static leakage analysis was performed using the \texttt{leakage-analysis} tool~\citep{yang2022dataleak}. This tool finds potential relations between the variables and outputs lines of code, causing potential data leakage. Leakage analysis is performed separately from the benchmark pipeline. Static leakage analysis finds potential leakage in $<$10\% of 
submissions in \texttt{MODULAR\_PREDICT} format, with 4\% remaining 
after excluding false positives in \texttt{train}-only and 
single-argument functions (Appendix~\ref{app:failures} and 
Figure~\ref{fig:leak_data}).

\subsection{Multilingual Expansion}
\label{sec:translating_data}

To obtain a multilingual corpus, we translate the \textit{domain}, 
\textit{description}, and \textit{data card} fields into target languages 
(Appendix~\ref{app:trans_prompt}). Other fields do not require translation 
since they convey universally recognized entities. Following 
\citet{jiao2023chatgptgoodtranslatoryes}, we use GPT-4o over commercial 
translators such as Google Translate and DeepL, as it matches or exceeds 
their quality for most language pairs.

After automatic translation, texts undergo manual validation by bilingual 
annotators with ML experience and at least a bachelor's degree in a 
computer-science--related field. Annotators are researchers from the authors' 
current and former affiliated institutions across multiple countries, ensuring 
authentic native-speaker validation; all participated voluntarily as academics 
collaborators and received co-authorship or acknowledgment commensurate with 
their contributions. We employ single-annotator validation per language, as 
GPT-4o's demonstrated translation quality means that annotators primarily perform 
error detection rather than de novo translation.

To structure validation, each annotator has received three Google Forms (one for 
each translated field). Each form presents the English source, the translation, 
and a question assessing whether the text sounds natural and preserves the 
original meaning. If the answer is ``No,'' the annotator has supplied a corrected 
version (Appendix~\ref{app:form_example}).

The final benchmark includes translations in Arabic, Belarusian, Chinese, 
French, Italian, Japanese, Kazakh, Polish, Romanian, Russian, Spanish, 
Turkish and Ukrainian, spanning diverse language families (Romance, Slavic, 
Semitic, Turkic, Sino-Tibetan, Japonic), writing systems (Latin, Cyrillic, 
Arabic, CJK), and resource levels. Language selection is constrained by the 
availability of qualified annotators. Translation quality is verified via 
back-translation BLEU (mean $>$0.35 for all language-field combinations) 
and inter-annotator agreement on a subset of 15 competitions (Gwet's AC1~\citep{Gwet2008Computing} $>$0.80 for meaning preservation across all languages; see Appendix~\ref{app:validation_results}).

\subsection{Evaluation}
\label{sec:evaluation}

Evaluating ML code generation across heterogeneous competitions requires careful metric design. Competitions use diverse evaluation metrics (e.g., RMSE, log-loss, F1-score, AUC) with incomparable scales and directionality. We address this through normalized performance measurement.

For each competition $p$, we establish two complementary in-benchmark baselines. The primary baseline uses LLM-assisted manual development: we iteratively prompt Gemini 3 Pro for solution approaches, implement standard ML pipelines (e.g., LightGBM with basic feature engineering), and validate on the 20\% validation split. The baseline score $b_p$ represents competent, manually-validated ML practice that handles heterogeneous task types uniformly. As a fully reproducible algorithmic reference point, we additionally provide an AutoGluon~\citep{agtabular, tang2024autogluon} 
baseline. We prefer AutoGluon over Auto-sklearn because it natively supports all data types in ML$^2$B (tabular, text, and image). 

For a model with score $s_p$ on competition $p$, we compute the 
performance ratio:
\begin{equation}
    r_p = \frac{e_p}{e_{b,p}}
\end{equation}
where $e_p$ and $e_{b,p}$ are the error rates of the model and baseline, 
respectively, defined as $e = 1 - s$ for higher-better metrics (e.g., 
F1, AUC) and $e = s$ for lower-better metrics (e.g., RMSE, log-loss). 
Thus, $r_p < 1$ indicates better-than-baseline performance, $r_p = 1$ 
parity, and $r_p > 1$ worse performance.

Following Dolan and Moré \citep{dolan2004benchmarking}, we define the performance profile as the cumulative distribution of performance ratios:
\begin{equation}
    \rho(\tau) = \frac{1}{N} \sum_{p=1}^{N} \mathbb{I}[r_p \leq \tau]
\end{equation}
representing the fraction of tasks where the model achieves performance within factor $\tau$ of baseline.

The standard Dolan-Moré formulation integrates $\rho(\tau)$ over $\tau \geq 1$ to measure worse-than-best performance. In our setting, where $r_p < 1$ indicates \emph{better}-than-baseline performance, we adapt the integration domain to $\tau \in [0, 1]$, yielding the standard area under the profile:
\begin{equation}
    \text{AUP}_{\text{std}} = \int_0^1 \rho(\tau) \, d\tau.
\end{equation}

This formulation emphasizes the region $r_p \leq 1$ (baseline-meeting or better performance). To emphasize different performance regimes, we use weighted AUP: 
\begin{equation}
    \text{AUP}_{w} = \frac{\int_0^1 \rho(\tau) \cdot w(\tau) \, d\tau}{\int_0^1 w(\tau) \, d\tau}.
\end{equation}
We report $\text{AUP}_{-1/\ln\tau}$ with $w(\tau) = -\frac{1}{\ln \tau}$, which emphasizes near-baseline distinctions ($r_p \approx 1$) and best distinguishes models that consistently meet baseline from those that occasionally succeed (formal definitions and numerical implementation details in Appendix~\ref{app:metrics}).

\section{Experiments and Results}

We evaluate three LLMs: GPT-4.1-mini, GPT-OSS-120b, and 
Gemini-2.5-Flash, using two agent frameworks: AIDE~\citep{jiang2025aide} and ReAct~\citep{yao2022react}. 
AIDE experiments use the standard configuration during code generation. For GPT-OSS-120b, we additionally evaluate a modified single-agent ReAct framework equipped with standard file system interaction capabilities (read, write, create) and the ability to execute Python and Bash commands. While the base ReAct agent cannot solve complex benchmark tasks requiring strictly structured Python files with predefined validation signatures, we address this limitation through two enhancements: tool execution timeouts prevent system freezing during prolonged operations, and the agent constructs structured command execution plans ensuring complete pipeline execution before termination.

Each model-framework configuration is executed 3 times per task with a 3-hour wall-clock time limit per run, yielding 630 runs per configuration (15 competitions × 14 languages × 3 runs). If executed sequentially in a single process, this amounts to 1,890 computing hours per configuration. A complete benchmark evaluation across all 4 model-framework configurations totals 2,520 runs and 7,560 sequential computing hours, though this duration can be reduced proportionally through parallel execution. All code execution is conducted on consumer-grade hardware representative of typical research environments: a workstation equipped with 64GB RAM and a single NVIDIA GeForce RTX 2080 Ti GPU (12GB VRAM). The average cost per configuration run ranges from $6.36$ USD for GPT-OSS-120b to $38.48$ USD for Gemini-2.5-Flash.

AutoGluon produces valid baselines on 13 of 15 competitions; two 
failures (Multi-label Classification, DataLab Cup) exceed GPU memory. 
Both baselines use the same 80/20 train/validation split (random 
seed~42) as all model evaluations and are available on GitHub. 
Table~\ref{tab:baseline_robustness} confirms that AUP rankings and 
core findings are consistent across both normalizations.

\begin{table}[ht]
\caption{AUP$_{-1/\ln\tau}$ $\pm$ std / Pass Rate (\%) under both 
baseline normalizations on the ML$^2$B chosen set (15 competitions, 
14 languages, 3 runs). Pass rates are baseline-independent.}
\centering
\small
\begin{tabular}{lcc}
\hline  
\textbf{Model / Framework} & \textbf{LLM-assisted baseline} & \textbf{AutoGluon baseline} \\
\hline 
GPT-4.1-mini (AIDE)      & .768 $\pm$ .112 / 61 & .736 $\pm$ .106 / 61 \\
Gemini-2.5-Flash (AIDE)  & .732 $\pm$ .091 / 76 & .775 $\pm$ .075 / 76 \\
GPT-OSS-120b (AIDE)      & .630 $\pm$ .083 / 81 & .616 $\pm$ .068 / 81 \\
GPT-OSS-120b (ReAct)     & .705 $\pm$ .081 / 76 & .732 $\pm$ .094 / 76 \\
\hline  
\end{tabular}
\label{tab:baseline_robustness}
\end{table}

Table~\ref{tab:main_results} presents the main results across all configurations. GPT-4.1-mini achieves the highest average AUP (0.768 $\pm$ 0.112) but with the lowest average pass rate (61\%), showing strong ranking ability but difficulty with exact solutions. Gemini-2.5-Flash demonstrates more balanced performance with high AUP (0.732 $\pm$ 0.091) and the best average pass rate (76\%). Among open-source models, GPT-OSS-120b in the AIDE framework shows a competitive AUP (0.630 $\pm$ 0.083) with a strong pass rate (81\%), while the proposed ReAct framework achieves a comparable AUP (0.705 $\pm$ 0.081) with 76\% pass rate despite disabled internet access.

\begin{table}[ht!]
\caption{
Cross-lingual performance on the ML$^2$B subset. 
$\mathrm{AUP}_{-1/\ln\tau}$ is reported as mean 
; PR denotes Pass Rate (\%). Higher is better. GPT-4.1-mini, GPT-OSS-120b, and Gemini-2.5-Flash are evaluated inside the AIDE execution. The proposed ReAct framework with GPT-OSS-120b is evaluated with the disabled internet access and external retrievers. In total, 2{,}520 runs are executed.
}
\centering
\small
\begin{tabular}{lcccc}  
\toprule
\textbf{Language} 
& GPT-4.1-mini (AIDE) & Gemini-2.5-Flash (AIDE) & GPT-OSS-120b (AIDE) & GPT-OSS (ReAct) \\  
\multicolumn{5}{c}{$\mathrm{AUP}_{-1/\ln\tau}$ / PR} \\  
\midrule
English & .756 $\pm$ .083 / 60 & .689 $\pm$ .157 / 84 & .600 $\pm$ .000 / 87 & .644 $\pm$ .063 / 78 \\
French & .711 $\pm$ .083 / 62 & .756 $\pm$ .175 / 80 & .556 $\pm$ .137 / 87 & .733 $\pm$ .144 / 73 \\
Spanish & .822 $\pm$ .083 / 51 & .689 $\pm$ .031 / 82 & .578 $\pm$ .083 / 91 & .733 $\pm$ .054 / 67 \\
Italian & .756 $\pm$ .137 / 62 & .800 $\pm$ .054 / 78 & .533 $\pm$ .054 / 96 & .622 $\pm$ .063 / 78 \\
Chinese & .711 $\pm$ .083 / 69 & .689 $\pm$ .166 / 84 & .667 $\pm$ .144 / 82 & .711 $\pm$ .031 / 73 \\
\hline
Japanese & .711 $\pm$ .191 / 71 & .778 $\pm$ .137 / 73 & .644 $\pm$ .063 / 73 & .667 $\pm$ .109 / 84 \\
Russian & .844 $\pm$ .083 / 56 & .800 $\pm$ .094 / 78 & .622 $\pm$ .063 / 78 & .711 $\pm$ .113 / 84 \\
Polish & .867 $\pm$ .054 / 49 & .711 $\pm$ .083 / 67 & .711 $\pm$ .083 / 78 & .778 $\pm$ .083 / 69 \\
Turkish & .822 $\pm$ .083 / 62 & .756 $\pm$ .113 / 73 & .689 $\pm$ .113 / 78 & .644 $\pm$ .083 / 84 \\
Arabic & .800 $\pm$ .109 / 60 & .689 $\pm$ .083 / 73 & .689 $\pm$ .083 / 73 & .800 $\pm$ .054 / 64 \\
Ukrainian & .622 $\pm$ .191 / 62 & .667 $\pm$ .000 / 80 & .711 $\pm$ .083 / 67 & .711 $\pm$ .083 / 80 \\
\hline
Romanian & .711 $\pm$ .113 / 64 & .756 $\pm$ .083 / 67 & .622 $\pm$ .083 / 78 & .711 $\pm$ .113 / 71 \\
Kazakh & .867 $\pm$ .094 / 67 & .711 $\pm$ .031 / 73 & .578 $\pm$ .083 / 89 & .711 $\pm$ .113 / 67 \\
Belarusian & .756 $\pm$ .175 / 60 & .756 $\pm$ .063 / 73 & .622 $\pm$ .083 / 78 & .689 $\pm$ .031 / 84 \\
\hline
\textbf{Avg.} & .768 $\pm$ .112 / 61 & .732 $\pm$ .091 / 76 & .630 $\pm$ .083 / 81 & .705 $\pm$ .081 / 76 \\
\bottomrule
\end{tabular}
\label{tab:main_results}
\end{table}


As shown in Appendix~\ref{app:gap_analysis}, the performance gap 
relative to English contradicts the traditional resource-availability 
hierarchy. High-resource languages exhibit mixed results, while medium- 
and low-resource languages display highly task-dependent behavior: 
certain competitions show better performance than English across multiple 
languages, while others exhibit substantial gaps. The variability within 
resource groups often exceeds the variability between them, suggesting 
that competition-specific characteristics (problem complexity, domain 
terminology, and code patterns) dominate over language resource 
availability. Task-characteristic analysis ($N$=15) further shows that 
feature count ($\rho$=+0.50) and domain-specific terminology 
($\rho$=+0.44) are the strongest predictors of cross-lingual variance 
(Appendix~\ref{app:task_analysis}).


We identify five failure categories: execution errors, constraint
violations, preprocessing inconsistencies, function signature
modifications, and environment issues, with GPT-OSS more robust to
these than GPT-4.1-mini (detailed breakdown and code examples in
Appendix~\ref{app:failures}). Our ReAct agent consumes on average 475K
tokens per experiment with a consistently low output-to-input ratio
($0.09$).

Taken together, our results reveal three consistent patterns. First,
cross-lingual gaps are task-dependent rather than resource-dependent:
within-group variability exceeds between-group variability across
resource levels (Appendix~\ref{app:gap_analysis}), and task
characteristics, i.e., feature count ($\rho$=+0.50) and domain
terminology ($\rho$=+0.44), predict cross-lingual variance better than
language identity. Second, the primary bottleneck is pipeline
instruction-following rather than language comprehension
(Appendix~\ref{app:failures}). Third, no model dominates on both
quality and reliability: GPT-4.1-mini attains the highest AUP but the
lowest pass rate, revealing a quality--robustness trade-off that
aggregate scores obscure.


\section{Conclusion}
We introduce ML$^2$B, the first multilingual benchmark for evaluating 
LLMs on end-to-end ML pipeline generation, comprising 35 real Kaggle 
competitions translated into 14 languages (490 tasks across tabular, 
text, and image data), validated by native speakers. ML$^2$B integrates 
methodological advances for reproducible evaluation: private 
competitions to reduce data leakage, modular code submission to limit 
preprocessing leakage, and containerized execution, though modular 
submissions may retain residual data access if agents perform 
late-stage training, which remains a future direction. Our findings 
reveal that cross-lingual performance gaps are highly task-dependent, 
challenging the assumption that resource availability alone determines 
multilingual model capabilities.

\appendix
\section{Prompt Format Comparison}
\label{app:prompt_comparison}

We validate that structured prompt format aids model comprehension without enabling memorization-based shortcuts. Using DeepSeek-Reasoner within AIDE (3-hour runtime, 3 runs), we compare ML$^2$B structured prompts against MLE-Bench obfuscated prompts on 10 shared competitions. As shown in  Figure~\ref{fig:prompt_comparison}, structured prompts achieve better performance on 5 of 10 competitions and demonstrate perfect stability (3/3 pass rate) across all tasks, whereas the obfuscated format fails (2/3 pass rate) on 5 competitions.

\definecolor{teal}{RGB}{0, 128, 128}
\definecolor{orange}{RGB}{255, 127, 14}

\tikzset{
full/.style={opacity=1.0},
medium/.style={opacity=0.55},
low/.style={pattern=north east lines}
}

\definecolor{teal}{RGB}{0, 128, 128}
\definecolor{orange}{RGB}{255, 127, 14}

\begin{figure}[t]
\centering
\begin{tikzpicture}
    \tikzmath{
        \spacing=0.6;  
        int \i;
        \i=0;
        for \rpos in {10, 9, 8, 7, 6, 5, 4, 3, 2, 1} {
            \rtops{\i} = \rpos*\spacing-0.2;
            \rbottoms{\i} = \rpos*\spacing+0.2;
            \rmids{\i} = \rpos*\spacing;
            \i=\i+1;
        };
        \ylim = \rbottoms{0}+0.3; 
        \ylimg = \rbottoms{0}+0.2; 
        \ybottom = 0.2;  
        \legendmid = \ybottom-0.65;
        \legendtop = \ybottom-0.8;
        \legendbottom = \ybottom-0.5;
    }

    \draw[->] (-5,\ybottom) -- (6,\ybottom) node[right] {\small\%};
    \draw (0,\ybottom) -- (0,\ylim);
    
    \foreach \x in {-4,-3,-2,-1,1,2,3,4,5} {
        \draw[gray!30] (\x,\ybottom) -- (\x,\ylimg);
        \node[below, font=\tiny] at (\x,\ybottom) {\pgfmathparse{int(\x*10)}\pgfmathresult};
    }
    
    
    \fill[teal] (0,\rtops{0}) rectangle (4.99,\rbottoms{0});
    \node[left, font=\small] at (0,\rmids{0}) {TPS Dec 2021 ($\uparrow$)};
    \node[right, font=\small] at (4.99,\rmids{0}) {+49.9\%};
    
    \fill[orange] (-4.39,\rtops{1}) rectangle (0,\rbottoms{1});
    \node[left, font=\small] at (0,\rmids{1}) {LMSYS Arena * ($\downarrow$)};
    \node[left, font=\small] at (-4.39,\rmids{1}) {-43.9\%};
    
    \fill[orange] (-2.49,\rtops{2}) rectangle (0,\rbottoms{2});
    \node[left, font=\small] at (0,\rmids{2}) {Dog Breed ($\downarrow$)};
    \node[left, font=\small] at (-2.49,\rmids{2}) {-24.9\%};
    
    \fill[teal] (0,\rtops{3}) rectangle (0.96,\rbottoms{3});
    \node[left, font=\small] at (0,\rmids{3}) {Author ID ($\downarrow$)};
    \node[right, font=\small] at (0.96,\rmids{3}) {+9.6\%};
    
    \fill[orange] (-0.75,\rtops{4}) rectangle (0,\rbottoms{4});
    \node[left, font=\small] at (0,\rmids{4}) {Jigsaw Toxicity * ($\uparrow$)};
    \node[left, font=\small] at (-2.99,\rmids{4}) {-7.5\%};
    
    \fill[teal] (0,\rtops{5}) rectangle (0.65,\rbottoms{5});
    \node[left, font=\small] at (0,\rmids{5}) {Insults Detection * ($\uparrow$)};
    \node[right, font=\small] at (0.65,\rmids{5}) {+6.5\% $^*$};
    
    \fill[teal] (0,\rtops{6}) rectangle (0.55,\rbottoms{6});
    \node[left, font=\small] at (0,\rmids{6}) {Hotel ID * ($\uparrow$)};
    \node[right, font=\small] at (0.55,\rmids{6}) {+5.5\%};
    
    \fill[teal] (0,\rtops{7}) rectangle (0.48,\rbottoms{7});
    \node[left, font=\small] at (0,\rmids{7}) {TPS May 2022 ($\uparrow$)};
    \node[right, font=\small] at (0.48,\rmids{7}) {+4.8\%};
    
    \fill[orange] (-0.19,\rtops{8}) rectangle (0,\rbottoms{8});
    \node[left, font=\small] at (0,\rmids{8}) {Ventilator Pressure ($\downarrow$)};
    \node[left, font=\small] at (-3.3,\rmids{8}) {-3.9\%};
    
    \fill[orange] (-0.01,\rtops{9}) rectangle (0,\rbottoms{9});
    \node[left, font=\small] at (0,\rmids{9}) {Cactus ID * ($\uparrow$)};
    \node[right, font=\small] at (0.01,\rmids{9}) {0.0\%};
    
    \fill[teal] (1.8,\legendtop) rectangle (2.0,\legendbottom);
    \node[right, font=\small] at (2.0,\legendmid) {Structured better};
    \fill[orange] (4.7,\legendtop) rectangle (4.9,\legendbottom);
    \node[right, font=\small] at (4.9,\legendmid) {Obfuscated better};
    
\end{tikzpicture}

\caption{Relative improvement of ML$^2$B structured prompts over MLE-Bench 
obfuscated prompts on 10 shared competitions using DeepSeek-Reasoner within 
AIDE framework (3-hour runtime, 3 runs per configuration). Arrows denote 
metric direction ($\uparrow$~higher, $\downarrow$~lower is better). 
Failed runs penalized with worst-case scores (0.0 for $\uparrow$, 100.0 
for $\downarrow$). * marks the obfuscated format failures (2/3 pass rate).}
\label{fig:prompt_comparison}
\end{figure}

\section{Full Competition List}
The full list of competitions is presented in Table~\ref{tab:all_competitions}.

\begin{table*}[]
\caption{The full list of competitions included in ML$^2$B}
\centering
\small
\resizebox{\linewidth}{!}{
\begin{tabular}{lcccccc}
\hline
\textbf{comp name} & \textbf{Private/Public} & \textbf{Year} & \textbf{data type} & \textbf{Teams} & \textbf{Source} \\
\hline
\textit{\href{https://www.kaggle.com/competitions/widsdatathon2020}{WiDS Datathon 2020}} & Public & 2020 & tabular & 951 & CODE4ML \\
\textit{\href{https://www.kaggle.com/competitions/ieor242hw4}{IEOR 242 Spring 2020 HW 4}} & Public & 2020 & tabular & 72 & CODE4ML \\
\textit{\href{https://www.kaggle.com/competitions/explicit-content-detection}{Explicit content detection}} & Public & 2020 & text & 106 & CODE4ML \\
\textit{\href{https://www.kaggle.com/competitions/109-1-ntut-dl-app-hw1}{109-1 NTUT Building Deep Learning Applications HW1}} & Public & 2020 & image & 90 & CODE4ML \\
\textit{\href{https://www.kaggle.com/competitions/stat441datachallenge1}{UWaterloo STAT441/841 Data Challenge 1}} & Public & 2020 & tabular & 131 & CODE4ML \\
\textit{\href{https://www.kaggle.com/competitions/made-hw-2}{MADE HW-2}} & Public & 2020 & text & 258 & CODE4ML \\
\textit{\href{https://www.kaggle.com/competitions/2020-10-29}{Financial Engineering Competition (1/3)}} & Public & 2020 & tabular & 290 & CODE4ML \\
\textit{\href{https://www.kaggle.com/competitions/2020-11-05}{Financial Engineering Competition (2/3)}} & Public & 2020 & tabular & 268 & CODE4ML \\
\textit{\href{https://www.kaggle.com/competitions/2020-11-20}{Financial Engineering Competition (3/3)}} & Public & 2020 & tabular & 268 & CODE4ML \\
\textit{\href{https://www.kaggle.com/competitions/prml-data-contest-nov-2020}{PRML-Data Contest-Nov 2020}} & Public & 2020 – 2021 & tabular & 150 & CODE4ML \\
\textit{\href{https://www.kaggle.com/competitions/actuarial-loss-estimation}{Actuarial loss prediction}} & Public & 2020 – 2021 & tabular & 140 & CODE4ML \\
\textit{\href{https://www.kaggle.com/competitions/shaastra-wells-fargo-hackathon}{<She/Hacks> - Shaastra'21 and Wells Fargo}} & Public & 2021 & tabular & 90 & CODE4ML \\
\textit{\href{https://www.kaggle.com/competitions/ml2021spring-hw1}{ML2021Spring-hw1}} & Public & 2021 & tabular & 2032 & CODE4ML \\
\textit{\href{https://www.kaggle.com/competitions/2021-ml-w1p1}{2021-ML (Korean)}} & Public & 2021 & tabular & 87 & CODE4ML \\
\textit{\href{https://www.kaggle.com/competitions/classification-with-non-deep-classifiers}{SYDE 522 (Winter 2021)}} & Public & 2021 & tabular & 130 & CODE4ML \\
\textit{\href{https://www.kaggle.com/competitions/tabular-playground-series-jul-2021}{Tabular Playground Series - Jul 2021}} & Public & 2021 & tabular & 1293 & CODE4ML \\
\textit{\href{https://www.kaggle.com/competitions/tabular-playground-series-aug-2021}{Tabular Playground Series - Aug 2021}} & Public & 2021 & tabular & 1753 & CODE4ML \\
\textit{\href{https://www.kaggle.com/competitions/classify-leaves}{Classify Leaves}} & Public & 2021 & image & 165 & CODE4ML \\
\textit{\href{https://www.kaggle.com/competitions/ventilator-pressure-prediction}{Google Brain - Ventilator Pressure Prediction}} & Public & 2021 & tabular & 2605 & CODE4ML \\
\textit{\href{https://www.kaggle.com/competitions/porto-seguro-data-challenge}{Porto Seguro Data Challenge}} & Public & 2021 & tabular & 174 & CODE4ML \\
\textit{\href{https://www.kaggle.com/competitions/crime-learn}{Crime Learn}} & Public & 2021 & tabular & 96 & CODE4ML \\
\textit{\href{https://www.kaggle.com/competitions/playground-series-s3e2}{Binary Classification with a Tabular Stroke Prediction Dataset}} & Public & 2023 & tabular & 770 & CODE4ML \\
\textit{\href{https://www.kaggle.com/competitions/ece460j-fall24/}{Multi-Class Prediction of Cirrhosis Outcomes}} & Private & 2024 & tabular & 108 & CODE4ML \\
\textit{\href{https://www.kaggle.com/competitions/playground-series-s5e6/data}{Predicting Optimal Fertilizers}} & Public & 2025 & tabular & 2648 & Kaggle \\
\textit{\href{https://www.kaggle.com/competitions/playground-series-s5e3/overview }{Binary Prediction with a Rainfall Dataset}} & Public & 2025 & tabular & 4381 & Kaggle \\
\textit{\href{https://www.kaggle.com/competitions/sheep-classification-challenge-2025}{Eid Al-Adha 2025: Sheep Classification Challenge}} & Public & 2025 & image & 355 & Kaggle \\
\textit{\href{https://www.kaggle.com/competitions/alfa-university-income-prediction}{Alfa University income prediction}} & Private & 2024 & tabular & 8 & Kaggle \\
\textit{\href{https://www.kaggle.com/competitions/2024-datalab-cup1/}{2024 DataLab Cup1}} & Private & 2024 & text & 108 & Kaggle \\
\textit{\href{https://www.kaggle.com/competitions/thapar-summer-school-2025-hack-iii/}{Thapar Summer School 2025 | Hack-III}} & Private & 2025 & tabular & 110 & Kaggle \\
\textit{\href{https://www.kaggle.com/competitions/rutgers-data101-fall2022-assignment-12/}{Rutgers Data101 Fall2022 Assignment 12}} & Private & 2022 & tabular & 162 & Kaggle \\
\textit{\href{https://www.kaggle.com/competitions/cs-506-fall-2025-technical-midterm }{CS 506 Fall 2025 Technical Midterm}} & Private & 2025 & tabular & 143 & Kaggle \\
\textit{\href{https://www.kaggle.com/competitions/car-becho-paisa-paao/overview }{Car Becho Paisa Paao}} & Private & 2025 & tabular & 302 & Kaggle \\
\textit{\href{https://www.kaggle.com/competitions/itmo-flat-price-prediction-2024}{ITMO Flat price prediction 2024}} & Private & 2024 – 2025 & tabular & 127 & Kaggle \\
\textit{\href{https://www.kaggle.com/competitions/multi-label-classification-competition-2025}{Multi-label Classification Competition 2025}} & Private & 2025 & image & 201 & Kaggle \\
\textit{\href{https://www.kaggle.com/competitions/ift-6390-ift-3395-beer-quality-prediction}{IFT6390-IFT3395: Beer Quality Prediction}} & Private & 2025 & tabular & 192 & Kaggle \\
\hline
\end{tabular}
}

\label{tab:all_competitions}
\end{table*}
\label{app:competition_list}

\section{Domain Subjects Gathering}

\label{app:domain}

To assign a suitable domain to every competition in our benchmark we have decided on a following process. Firstly, the prompt containing the information about the problem (Figure~\ref{fig:domain_prompt}) was given to the GPT-3.5-turbo model to derive domain tag for each competition. The next step was manual evaluation and necessary polishing of LLM-assigned domains in order to ensure their accuracy and suitability. Finally, 9 concise domains are obtained: media studies, social studies, environmental studies, urban studies, finance, marketing, healthcare and medical, manufacturing and quality, synthetic data. The number of competitions in each domain is presented in Figure~\ref{fig:domains_distr_compare}.

\begin{figure}[h!]
\centering
\begin{tcolorbox}
[colback=cyan!10!green!10, colframe=cyan!50!black]
You are given competition name, data card and description of Kaggle competition. 
You need to identify the domain that the task belongs to in the given competition.\\[0.25em]
\textbf{Competition name:} Crime\_Learn\\[0.25em]
\textbf{Description:} Develop a predictive model to estimate the rate of violent crimes per population in a given area based on specific features. The input consists of two datasets, one for training and one for testing, with the target variable being 'ViolentCrimesPerPop'.\\[0.25em] 
\textbf{Data card:} In this competition you will use the sample US crime data for predicting 'ViolentCrimesPerPop'. train.csv – the training dataset.\\[0.25em]
\end{tcolorbox}
\caption{Example of the prompt used to derive competition domain}
\label{fig:domain_prompt}
\end{figure}

\begin{figure}[h!]
\centering
\begin{tikzpicture}
\begin{axis}[
    ybar,
    width=\columnwidth,
    height=5cm,
    bar width=6pt,
    ymin=0,
    ymax=8,
    ylabel={Count},
    symbolic x coords={
    finance,healthcare and medical,environmental studies,
    media studies,synthetic data,social studies,
    urban studies,marketing,manufacturing and quality
    },
    xtick=data,
    xticklabel style={
        rotate=45,
        anchor=east,
        font=\footnotesize
    },
    ymajorgrids=true,
    grid style={gray!30},
    legend style={
        at={(0.5,1.05)},
        anchor=south,
        legend columns=2,
    },
    enlarge x limits=0.15,
]
\addplot[fill=teal] coordinates {
    (media studies,4) (marketing,2) (environmental studies,6)
    (finance,7) (healthcare and medical,6) (manufacturing and quality,1)
    (social studies,3) (synthetic data,4) (urban studies,2)
};

\addplot[fill=orange] coordinates {
    (media studies,3) (marketing,0) (environmental studies,1)
    (finance,4) (healthcare and medical,3) (manufacturing and quality,1)
    (social studies,0) (synthetic data,2) (urban studies,1)
};

\legend{Full set \hspace{0.3cm} , Sub set}
\end{axis}
\end{tikzpicture}
\caption{Domain distribution comparison on full and sub-set}
\label{fig:domains_distr_compare}
\end{figure}

\section{Prompt Templates}
\label{app:prompt_example_competition}

Prompts consist of three sections: (1) benchmark instructions, 
(2) competition-specific task description, and (3) submission constraints. 
Table~\ref{tab:prompt_sections} summarizes each section's content, while 
Table~\ref{tab:signatures} shows the four supported function signature variants.

\begin{table}[t]
\caption{Prompt template structure.}
\centering
\small
\setlength{\tabcolsep}{4pt}
\begin{tabular}{p{5cm}p{10cm}}
\toprule
\textbf{Section} & \textbf{Content} \\
\midrule
Benchmark instructions & 
Submission path requirements, train/test split clarification, programming language constraints, anti-plagiarism policy \\

Task description & 
ML task objective, domain, target metric, data schema with column descriptions (translated to target language) \\

Submission constraints & 
Required function signatures, single-file requirement, no global variables, self-contained code \\
\bottomrule
\end{tabular}
\label{tab:prompt_sections}
\end{table}

\begin{table}[t]
\caption{Supported function signature variants.}
\centering
\small
\setlength{\tabcolsep}{4pt}
\begin{tabular}{p{5cm}p{10cm}}
\toprule
\textbf{Type} & \textbf{Signature} \\
\midrule
Monolithic (extended) & \texttt{train\_and\_predict(...)} \\

Monolithic (TypedDict) & \texttt{train\_and\_predict(X\_train, X\_val)} \\

Modular (extended) & \texttt{train()}, \texttt{prepare\_val()}, \texttt{predict()} \\

Modular (TypedDict) & Same with TypedDict \\
\bottomrule
\end{tabular}
\label{tab:signatures}
\end{table}

\paragraph{Function Signature Variants.}
We support four signature variants based on two dimensions as shown in 
Table~\ref{tab:signatures}. Full prompt templates with complete examples are available in our 
code repository.\footnote{\url{https://github.com/enaix/ml2b}}

\noindent\textbf{Monolithic} variant requires a single function:
\begin{lstlisting}[language=Python, basicstyle=\small\ttfamily]
def train_and_predict(X_train: dict, X_val: dict) -> np.ndarray:
    """Train model and return predictions."""
    ...
\end{lstlisting}

\noindent\textbf{Modular} variant separates training and inference:
\begin{lstlisting}[language=Python, basicstyle=\small\ttfamily]
def train(X_train: dict) -> Any: ...
def prepare_val(train_output: Any, X_val: dict) -> Any: ...
def predict(train_output: Any, val_output: Any) -> np.ndarray: ...
\end{lstlisting}

\section{Failure Pattern Analysis}
\label{app:failures}

Out of 2{,}520 submissions in the \texttt{MODULAR\_PREDICT} format, 
$<$10\% contain potential data leakage according to static analysis. 
In 81 submissions, leakage is detected in the \texttt{train} function, 
which does not operate on prediction data; in 33 cases it appears in 
trivial single-argument functions --- both are false positives. This 
leaves 4\% of submissions with potential actual leakage. Residual 
leakage may still occur if agents perform model training in later 
pipeline stages where both training and prediction data are 
theoretically accessible.

\begin{figure*}[t!]
\centering
\includegraphics[trim=0 0 0 0, clip, width=1\textwidth]{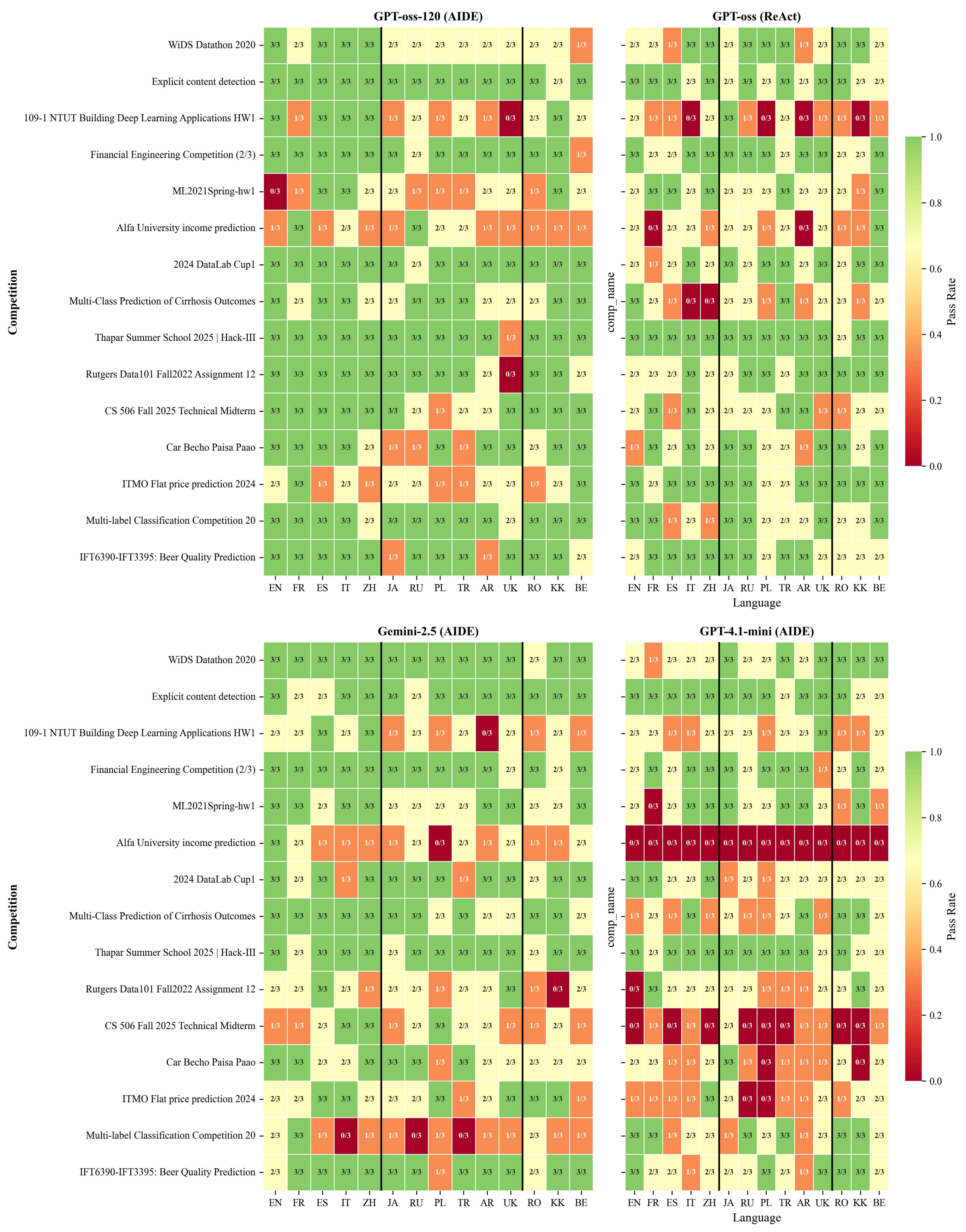}
\caption{Pass rate across 14 languages and 15 competitions}
\label{fig:pass_rate}
\end{figure*}

Table~\ref{tab:failure_patterns} categorizes the systematic failures 
observed across our experiments (pass rates across all languages and 
competitions are shown in Figure~\ref{fig:pass_rate}).

\begin{table}[h]
\caption{Detailed failure patterns observed across ML$^2$B experiments.}
\centering

\small
\begin{tabular}{p{5cm}p{10cm}}
\toprule
\textbf{Category} & \textbf{Description} \\
\midrule
Execution errors & 
Absence of \texttt{if \_\_name\_\_ == "\_\_main\_\_"} block prevented importing target functions from submission file \\
\addlinespace
Data handling & 
Agents directly load data files within training functions instead of using function parameters, violating the predefined function signatures that expect data as arguments. \\
\addlinespace
Preprocessing mismatch & 
Inconsistent feature engineering between training and validation sets, 
causing dimension or type errors \\
\addlinespace
Signature violations & 
Modified function signatures despite explicit instructions requiring 
exact formats for grader compatibility \\
\addlinespace
Global dependencies & 
Initialization code placed outside function definitions, violating 
self-contained code requirements \\
\addlinespace
Environment issues & 
Use of deprecated API calls, non-existent library functions, or 
incorrect package versions \\
\bottomrule
\end{tabular}
\label{tab:failure_patterns}
\end{table}

Notably, these failures occurred more frequently with GPT-4.1-mini than 
with GPT-oss or Gemini-2.5-flash variants, suggesting that larger or more recent models may 
have better instruction-following capabilities for constrained code 
generation tasks. 

Listings~\ref{lst:valid} and~\ref{lst:failed_code},~\ref{lst:failed_code2} provide the examples of valid and two failed GPT-4.1-mini-generated codes, correspondingly.

\label{app:code_example}
\begin{lstlisting}[
floatplacement=H,
    caption={Valid submission with proper modular structure and LightGBM pipeline.},
    label={lst:valid},
    basicstyle=\tiny\ttfamily,
    numbers=left,
    numberstyle=\tiny,
    stepnumber=1,
    numbersep=6pt
]
import pandas as pd
import numpy as np
from typing import Any
from sklearn.model_selection import GroupKFold
from sklearn.metrics import mean_absolute_error
import lightgbm as lgb
def create_features(df):
    # Basic features
    df["R"] = df["R"].astype("category")
    df["C"] = df["C"].astype("category")
    # Lag features for u_in and u_out to capture time series
    df["u_in_lag1"] = df.groupby("breath_id")["u_in"].shift(1).fillna(0)
    df["u_in_lag2"] = df.groupby("breath_id")["u_in"].shift(2).fillna(0)
    df["u_out_lag1"] = df.groupby("breath_id")["u_out"].shift(1).fillna(0)
    df["u_out_lag2"] = df.groupby("breath_id")["u_out"].shift(2).fillna(0)
    # Time step as feature
    df["time_step"] = df["time_step"].astype(np.float32)
    # We can also add cumulative sum of u_in per breath to capture progression
    df["u_in_cumsum"] = df.groupby("breath_id")["u_in"].cumsum()
    # Drop columns not used for training
    return df
def train(X_train: pd.DataFrame, y_train: pd.DataFrame) -> Any:
    X_train = create_features(X_train)
    features = [
        "R",
        "C",
        "time_step",
        "u_in",
        "u_out",
        "u_in_lag1",
        "u_in_lag2",
        "u_out_lag1",
        "u_out_lag2",
        "u_in_cumsum",
    ]
    categorical_features = ["R", "C", "u_out", "u_out_lag1", "u_out_lag2"]
    # Prepare dataset for LightGBM
    lgb_train = lgb.Dataset(
        X_train[features],
        label=y_train["pressure"],
        categorical_feature=categorical_features,
        free_raw_data=False,
    )
    params = {
        "objective": "regression",
        "metric": "mae",
        "boosting_type": "gbdt",
        "learning_rate": 0.1,
        "num_leaves": 31,
        "max_depth": -1,
        "verbose": -1,
        "seed": 42,
        "feature_pre_filter": False,
    }
    model = lgb.train(params, lgb_train, num_boost_round=100)
    return (model, features, categorical_features)
def prepare_val(X_val: pd.DataFrame, train_output: Any) -> Any:
    model, features, categorical_features = train_output
    X_val = create_features(X_val)
    return X_val[features]
def predict(train_output: Any, prepare_val_output: Any) -> np.array:
    model, features, categorical_features = train_output
    X_val = prepare_val_output
    preds = model.predict(X_val, num_iteration=model.best_iteration)
    return preds
def run(X_train: pd.DataFrame, y_train: pd.DataFrame, X_val: pd.DataFrame) -> np.array:
    train_output = train(X_train, y_train)
    X_val_processed = prepare_val(X_val, train_output)
    preds = predict(train_output, X_val_processed)
    return preds
\end{lstlisting}

\begin{lstlisting}[
floatplacement=H,
    caption={Failed submission illustrating API version incompatibility (line 55). The agent attempts to use \texttt{early\_stopping\_rounds} parameter in \texttt{lgb.train()}, which was deprecated in LightGBM 3.0.0 and removed in 4.0.0. This demonstrates a systematic limitation: agents do not verify library version compatibility or adapt their code when receiving error feedback about deprecated APIs.},
label={lst:failed_code},
    basicstyle=\tiny\ttfamily,
    numbers=left,
    numberstyle=\tiny,
    stepnumber=1,
    numbersep=6pt
]
import pandas as pd
import numpy as np
from typing import Any
from sklearn.model_selection import GroupKFold
from sklearn.metrics import mean_absolute_error
import lightgbm as lgb
import os
def feature_engineering(df: pd.DataFrame) -> pd.DataFrame:
    # Basic features
    df["u_in_cumsum"] = df.groupby("breath_id")["u_in"].cumsum()
    df["u_in_lag1"] = df.groupby("breath_id")["u_in"].shift(1).fillna(0)
    df["u_in_lag2"] = df.groupby("breath_id")["u_in"].shift(2).fillna(0)
    df["u_out_lag1"] = df.groupby("breath_id")["u_out"].shift(1).fillna(0)
    df["time_step_diff"] = df.groupby("breath_id")["time_step"].diff().fillna(0)
    # Interaction features
    df["R*C"] = df["R"] * df["C"]
    df["R*u_in"] = df["R"] * df["u_in"]
    df["C*u_in"] = df["C"] * df["u_in"]
    return df
def train(X_train: pd.DataFrame, y_train: pd.DataFrame) -> Any:
    X_train = feature_engineering(X_train)
    features = [
        "R",
        "C",
        "time_step",
        "u_in",
        "u_out",
        "u_in_cumsum",
        "u_in_lag1",
        "u_in_lag2",
        "u_out_lag1",
        "time_step_diff",
        "R*C",
        "R*u_in",
        "C*u_in",
    ]
    X_train = X_train[features]
    y = y_train["pressure"].values
    # GroupKFold on breath_id to avoid leakage
    # We will train a single model on full data here for simplicity
    params = {
        "objective": "regression",
        "metric": "mae",
        "boosting_type": "gbdt",
        "learning_rate": 0.1,
        "num_leaves": 64,
        "feature_fraction": 0.8,
        "bagging_fraction": 0.8,
        "bagging_freq": 5,
        "verbose": -1,
        "seed": 42,
        "n_jobs": -1,
    }
    lgb_train = lgb.Dataset(X_train, y)
    model = lgb.train(
        params,
        lgb_train,
        num_boost_round=1000,
        valid_sets=[lgb_train],
        early_stopping_rounds=50, # <-- Unsupported since version 4.0.0
        verbose_eval=100,
    )
    return (model, features)
def prepare_val(X_val: pd.DataFrame, train_output: Any) -> Any:
    model, features = train_output
    X_val = feature_engineering(X_val)
    X_val = X_val[features]
    return X_val
def predict(train_output: Any, prepare_val_output: Any) -> np.array:
    model, _ = train_output
    X_val = prepare_val_output
    preds = model.predict(X_val, num_iteration=model.best_iteration)
    return preds
def run(X_train: pd.DataFrame, y_train: pd.DataFrame, X_val: pd.DataFrame) -> np.array:
    train_output = train(X_train, y_train)
    prepared_val = prepare_val(X_val, train_output)
    preds = predict(train_output, prepared_val)
    return preds
\end{lstlisting}

\begin{lstlisting}[
    caption={Data leakage through target-dependent feature engineering. In train() (line 35), the agent concatenates input features with the target variable and applies add\_features(), which generates lag features based on the target column pressure (lines 13–14). Reusing this feature engineering pipeline on the validation set (lines 62, 70), where the target is not available, results in a missing-column error.},
label={lst:failed_code2},
    basicstyle=\tiny\ttfamily,
    numbers=left,
    numberstyle=\tiny,
    stepnumber=1,
    numbersep=6pt
]
import pandas as pd
import numpy as np
from typing import Any, Tuple
from sklearn.model_selection import train_test_split
from sklearn.metrics import mean_absolute_error
import lightgbm as lgb
def add_features(df: pd.DataFrame) -> pd.DataFrame: # <-- using target column presure
    # Sort by breath_id and time_step for lag features
    df = df.sort_values(["breath_id", "time_step"]).reset_index(drop=True)
    # Lag features for u_in and pressure
    df["u_in_lag1"] = df.groupby("breath_id")["u_in"].shift(1).fillna(0)
    df["u_in_lag2"] = df.groupby("breath_id")["u_in"].shift(2).fillna(0)
    df["pressure_lag1"] = df.groupby("breath_id")["pressure"].shift(1).fillna(0)
    df["pressure_lag2"] = df.groupby("breath_id")["pressure"].shift(2).fillna(0)
    # Rolling mean of u_in over last 3 steps
    df["u_in_roll3"] = (
        df.groupby("breath_id")["u_in"]
        .rolling(3, min_periods=1)
        .mean()
        .reset_index(0, drop=True)
    )
    # Rolling std dev of u_in over last 3 steps
    df["u_in_roll3_std"] = (
        df.groupby("breath_id")["u_in"]
        .rolling(3, min_periods=1)
        .std()
        .reset_index(0, drop=True)
        .fillna(0)
    )
    # Time step squared to capture nonlinear trend
    df["time_step2"] = df["time_step"] ** 2
    return df
def train(X_train: pd.DataFrame, y_train: pd.DataFrame) -> Any:
    # Add features to training data
    X_train = add_features(pd.concat([X_train, y_train], axis=1))
    # Target
    y = X_train["pressure"].values
    # Drop columns not used as features
    drop_cols = ["id", "breath_id", "pressure"]
    features = [c for c in X_train.columns if c not in drop_cols]
    X = X_train[features]
    # LightGBM dataset
    lgb_train = lgb.Dataset(X, label=y)
    params = {
        "objective": "regression",
        "metric": "mae",
        "verbosity": -1,
        "boosting_type": "gbdt",
        "learning_rate": 0.1,
        "num_leaves": 31,
        "feature_fraction": 0.9,
        "bagging_fraction": 0.9,
        "bagging_freq": 1,
        "seed": 42,
        "n_jobs": -1,
    }
    model = lgb.train(params, lgb_train, num_boost_round=300)
    return (model, features)
def prepare_val(X_val: pd.DataFrame, train_output: Any) -> Any:
    model, features = train_output
    # Add features to validation data (pressure not available, fill with 0)
    X_val = add_features(X_val.copy()) # <-- pressure target column not presented
    # For lags of pressure which we don't have in val, fill with 0
    X_val["pressure"] = 0.0
    X_val["pressure_lag1"] = 0.0
    X_val["pressure_lag2"] = 0.0
    X_val = add_features(X_val)
    return X_val[features]
def predict(train_output: Any, prepare_val_output: Any) -> np.array:
    model, features = train_output
    X_val = prepare_val_output
    preds = model.predict(X_val)
    return preds
def run(X_train: pd.DataFrame, y_train: pd.DataFrame, X_val: pd.DataFrame) -> np.array:
    train_output = train(X_train, y_train)
    val_features = prepare_val(X_val, train_output)
    preds = predict(train_output, val_features)
    return preds
\end{lstlisting}

\FloatBarrier

\section{Translation Prompt}
\label{app:trans_prompt}
Since some fields include imperatives (e.g., \textit{Develop a model}, \textit{Create an agent}), it has to be defined explicitly in a prompt (Figure \ref{fig:prompt_example}) to use imperative mood, otherwise the model have translated English imperatives, which have the same form as verbs not in imperative mood, mainly as infinitives.

\begin{figure}[h!]
\centering
\begin{tcolorbox}[colback=cyan!10!green!10, colframe=cyan!50!black, title=Prompt example, sharp corners]
Translate text into \textbf{\{target\_language\}}. Infinitive forms that stand apart, if any, should be translated as the imperative mood: \textbf{\{text\}}
\end{tcolorbox}
\caption{Example of the prompt used in translation experiments.}
\label{fig:prompt_example}
\end{figure}

\label{app:domain_prompt}


\section{Form Example}
\label{app:form_example}

Figure~\ref{fig:translation-question} illustrates an example of one question block in a form, which asks a native speaker of Romanian to validate the translation of the competition description.

\begin{figure}[h!]
\centering
\begin{tcolorbox}[colback=cyan!10!green!10, colframe=cyan!50!black, title=Translation Check]
\textbf{Translated version:} \\[0.5em] 
Dezvoltați un model predictiv pentru a anticipa probabilitatea mortalității în spital pentru pacienți. Seturile de date includ diverse caracteristici legate de pacienți la momentul internării în spital. Obiectivul este de a prezice cu acuratețe probabilitatea mortalității în spital pentru fiecare pacient din setul de testare.

\textbf{Original version:} \\[0.5em]
Develop a predictive model to forecast the likelihood of hospital mortality for patients. The datasets include various features related to the patients upon hospital admission. The objective is to predict the probability of hospital mortality for each patient in the test set accurately.

\textbf{Does the translated text (1) sound native and (2) convey the same meaning as the original text?}

\begin{itemize}
  \item YES and YES
  \item NO and YES
  \item YES and NO
  \item NO and NO
\end{itemize}

If there is at least one NO in the answer, please suggest your own version:

\end{tcolorbox}
\caption{Example of question block in Google Form for Romanian language}
\label{fig:translation-question}
\end{figure}

\section{Validation Results}
\label{app:validation_results}

As it has been stated before, for each language we considered the translations from a single annotator. To prove the reliability and the quality of translations, we have used back-translation into English and have assessed quality using BLEU metric.

Using the same GPT-4o model and identical prompts, we have executed three independent translation runs and collected the resulting English back-translations. We then have computed BLEU scores between the original texts in English and their back-translated versions. Finally, we have estimated bootstrap confidence intervals for each language and text type.

\citet{lavie-2010-evaluating} claims that a BLEU score in the range [0.4, 0.5] can be a sign of high-quality translation. As shown in Figure~\ref{fig:bleu-confidence}, the mean BLEU scores across the three runs exceed 0.35 for all languages. Moreover, the intervals for descriptions and data cards are relatively narrow, suggesting stable and accurate translations.

For domain-level texts, the confidence intervals are notably wider. This is expected since BLEU is highly sensitive to short texts, and GPT-4o often produces less consistent translations for very short inputs lacking contextual cues. Overall, these results support our claim that translations from a single native speaker are sufficiently reliable for our study.

\begin{figure}[t]
\centering
\begin{tikzpicture}
\begin{axis}[
    width=\columnwidth,
    height=5.8cm,
    ylabel={BLEU},
    ylabel style={font=\footnotesize, at={(axis description cs:-0.06,.5)}},
    symbolic x coords={ar,be,zh,fr,it,ja,kk,pl,ro,ru,es,tr,uk},
    xtick=data,
    xticklabels={
    Arabic,
    Belarusian,
    Chinese,
    French,
    Italian,
    Japanese,
    Kazakh,
    Polish,
    Romanian,
    Russian,
    Spanish,
    Turkish,
    Ukrainian},
    xticklabel style={rotate=35,anchor=east,font=\small},
    yticklabel style={font=\tiny},
    ytick={0.3,0.4,0.5,0.6,0.7,0.8,0.9,1.0},
    ymin=0.28, ymax=1.02,
    legend style={
        at={(0.5,1.02)},
        anchor=south,
        legend columns=3,
        font=\scriptsize,
        column sep=3pt,
        draw=none,
    },
    ymajorgrids=true,
    grid style={dashed, gray!25},
    enlarge x limits=0.055,
    clip=false,
]

\addplot[
    only marks,
    mark=triangle*,
    mark size=2.2pt,
    color=blue!75!black,
    error bars/.cd,
    y dir=both,
    y explicit,
    error bar style={line width=0.6pt},
] coordinates {
    (ar, 0.4121) +- (0, 0.0493)
    (be, 0.4063) +- (0, 0.0480)
    (zh, 0.3996) +- (0, 0.0443)
    (fr, 0.5803) +- (0, 0.0482)
    (it, 0.5447) +- (0, 0.0554)
    (ja, 0.3741) +- (0, 0.0420)
    (kk, 0.3406) +- (0, 0.0428)
    (pl, 0.3681) +- (0, 0.0430)
    (ro, 0.4326) +- (0, 0.0437)
    (ru, 0.3594) +- (0, 0.0470)
    (es, 0.6453) +- (0, 0.0540)
    (tr, 0.4080) +- (0, 0.0461)
    (uk, 0.4443) +- (0, 0.0472)
};

\addplot[
    only marks,
    mark=*,
    mark size=2.2pt,
    color=orange!85!black,
    error bars/.cd,
    y dir=both,
    y explicit,
    error bar style={line width=0.6pt},
] coordinates {
    (ar, 0.5781) +- (0, 0.0416)
    (be, 0.5149) +- (0, 0.0382)
    (zh, 0.5524) +- (0, 0.0421)
    (fr, 0.6184) +- (0, 0.0406)
    (it, 0.6640) +- (0, 0.0398)
    (ja, 0.4862) +- (0, 0.0351)
    (kk, 0.4967) +- (0, 0.0426)
    (pl, 0.6000) +- (0, 0.0396)
    (ro, 0.6183) +- (0, 0.0510)
    (ru, 0.5389) +- (0, 0.0375)
    (es, 0.6661) +- (0, 0.0433)
    (tr, 0.5651) +- (0, 0.0398)
    (uk, 0.5639) +- (0, 0.0380)
};

\addplot[
    only marks,
    mark=square*,
    mark size=2.2pt,
    color=teal!85!black,
    error bars/.cd,
    y dir=both,
    y explicit,
    error bar style={line width=0.6pt},
] coordinates {
    (ar, 0.4387) +- (0, 0.0797)
    (be, 0.6026) +- (0, 0.0887)
    (zh, 0.7827) +- (0, 0.0674)
    (fr, 0.3934) +- (0, 0.0752)
    (it, 0.6258) +- (0, 0.0836)
    (ja, 0.7425) +- (0, 0.0791)
    (kk, 0.6448) +- (0, 0.0829)
    (pl, 0.6625) +- (0, 0.0818)
    (ro, 0.6175) +- (0, 0.0890)
    (ru, 0.6594) +- (0, 0.0770)
    (es, 0.4837) +- (0, 0.0795)
    (tr, 0.9203) +- (0, 0.0508)
    (uk, 0.9462) +- (0, 0.0427)
};

\legend{data\_card, description, domain}
\end{axis}
\end{tikzpicture}
\caption{BLEU scores with 95\% confidence intervals by language and text type.}
\label{fig:bleu-confidence}
\end{figure}

Table~\ref{tab:agreement_ac1} shows the inter-annotator agreement measured by Gwet’s AC1.

\label{app:agreement}

\begin{table}[ht]
\caption{Inter-annotator agreement measured by Gwet’s AC1 for nativeness and meaning preservation criteria across languages.}
\centering
\begin{tabular}{lcc}
\hline
\textbf{Language} & \textbf{Nativeness AC1} & \textbf{Meaning AC1} \\
\hline
Arabic     & 0.936 & 0.902 \\
Belarusian & 0.950 & 0.888 \\
Chinese    & 0.450 & 0.971 \\
French     & 0.723 & 0.899 \\
Italian    & 0.841 & 0.971 \\
Japanese   & 0.308 & 1.000 \\
Kazakh     & 0.359 & 1.000 \\
Polish     & 0.743 & 0.931 \\
Romanian   & 0.560 & 0.869 \\
Russian    & 0.876 & 0.971 \\
Spanish    & 0.773 & 0.822 \\
Turkish    & 0.876 & 1.000 \\
Ukrainian  & 0.538 & 1.000 \\
\hline
\end{tabular}
\label{tab:agreement_ac1}
\end{table}


\begin{figure}[h!]
\centering
\begin{tikzpicture}
\begin{axis}[
    ybar stacked,
    width=0.95\columnwidth,
    height=5cm,
    bar width=10pt,
    ymin=0,
    ymax=82,
    ylabel={Count},
    symbolic x coords={
        Arabic,Belarusian,Chinese,French,Italian,Japanese,Kazakh,
        Polish,Romanian,Russian,Spanish,Turkish,Ukrainian
    },
    xtick=data,
    xticklabel style={rotate=35,anchor=east,font=\small},
    ymajorgrids=true,
    grid style={gray!30},
    legend style={
    at={(0.5,1.15)},
    anchor=south,
    legend columns=-1,
    draw=none,
    font=\small
},
    nodes near coords,
    every node near coord/.append style={
        font=\scriptsize,
        color=white
    },
]
\addplot[fill=teal] coordinates {
(Arabic,63)
(Belarusian,73)
(Chinese,73)
(French,53)
(Italian,58)
(Japanese,64)
(Kazakh,47)
(Polish,74)
(Romanian,55)
(Russian,58)
(Spanish,57)
(Turkish,73)
(Ukrainian,74)
};

\addplot[fill=blue] coordinates {
(Arabic,6)
(Belarusian,2)
(Chinese,2)
(French,2)
(Italian,3)
(Japanese,0)
(Kazakh,3)
(Polish,2)
(Romanian,4)
(Russian,0)
(Spanish,1)
(Turkish,3)
(Ukrainian,4)
};

\addplot[fill=orange] coordinates {
(Arabic,5)
(Belarusian,3)
(Chinese,3)
(French,24)
(Italian,18)
(Japanese,15)
(Kazakh,24)
(Polish,2)
(Romanian,20)
(Russian,19)
(Spanish,18)
(Turkish,2)
(Ukrainian,1)
};

\addplot[fill=brown] coordinates {
(Arabic,5)
(Belarusian,1)
(Chinese,1)
(French,0)
(Italian,0)
(Japanese,0)
(Kazakh,5)
(Polish,1)
(Romanian,0)
(Russian,2)
(Spanish,3)
(Turkish,1)
(Ukrainian,0)
};
\legend{YES and YES, YES and NO, NO and YES, NO and NO}
\end{axis}
\end{tikzpicture}
\caption{Answer distribution across languages}
\label{fig:lang_bar}
\end{figure}

We also analyze the GPT-4 translation patterns. Prior work shows that GPT-4 produces more accurate and more lexically diverse translations than commercial systems such as Google Translate \citep{jiao2023chatgptgoodtranslatoryes}. Additionally, \citet{raunak2023gptsproduceliteraltranslations} note that GPT-family models tend to generate non-literal translations, including figurative renderings of idioms. Based on this, we anticipated many outputs that preserved meaning but lacked features that make them sound fully native.

\label{app:translating_patterns}
Figure~\ref{fig:lang_bar} indicates that nearly two-thirds of responses within each language are judged as both natural and semantically equivalent. The second most frequent pattern is NO and YES, reflecting translations that preserve meaning but lack native-like phrasing. This outcome is expected for LLMs, which tend to prioritize semantic adequacy over fine-grained stylistic and idiomatic conventions~\citep{Mahowald2024Dissociating}.

\section{AUP Metric Details}
\label{app:metrics}
Here we provide the details on the modified versions of AUP metric.


\paragraph{Weighted AUP with Emphasis on Strong Outperformance.}
\begin{equation}
    \text{AUP}_{-\ln\tau} = \int_0^1 \rho(\tau) \cdot (-\ln\tau) \, d\tau
\end{equation}
The weight $w(\tau) = -\ln\tau$ is strictly positive on $(0,1)$, with 
$\int_0^1 (-\ln\tau) d\tau = 1$, ensuring natural normalization. This 
metric is sensitive to models achieving $r_p \ll 1$.

\paragraph{Weighted AUP with Emphasis on Near-Baseline Performance.}
\begin{equation}
    \text{AUP}_{-1/\ln\tau} = \frac{\int_{\epsilon}^{1-\epsilon} \rho(\tau) \cdot \left(-\frac{1}{\ln\tau}\right) d\tau}{\int_{\epsilon}^{1-\epsilon} \left(-\frac{1}{\ln\tau}\right) d\tau}
\end{equation}
where $\epsilon = 10^{-8}$. This metric emphasizes $\tau \approx 1$, 
distinguishing models that reliably meet baseline from those slightly below.

\paragraph{Sensitivity Properties of Weighted AUP Metrics}

The weighting functions transform the standard AUP into complementary metrics with distinct sensitivity profiles. For $\mathrm{AUP}_{-\ln\tau}$, the weight $w(\tau) = -\ln\tau$ diverges as $\tau \to 0^+$, making the metric highly sensitive to models achieving $r_{p,s} \ll 1$—i.e., models that substantially outperform the baseline. This metric, therefore, highlights \textbf{excellence}, emphasizing large performance gains and breakthrough capabilities.

Conversely, for $\mathrm{AUP}_{-1/\ln\tau}$, the weight $w(\tau) = -1/\ln\tau$ diverges as $\tau \to 1^-$, amplifying differences in the critical region where models are only marginally better than the baseline ($r_{p,s} \approx 1$). This metric emphasizes \textbf{reliability}, rewarding consistent, modest improvements over the baseline.

\begin{figure}[ht!]
\begin{center}
\includegraphics[width=0.8\linewidth]{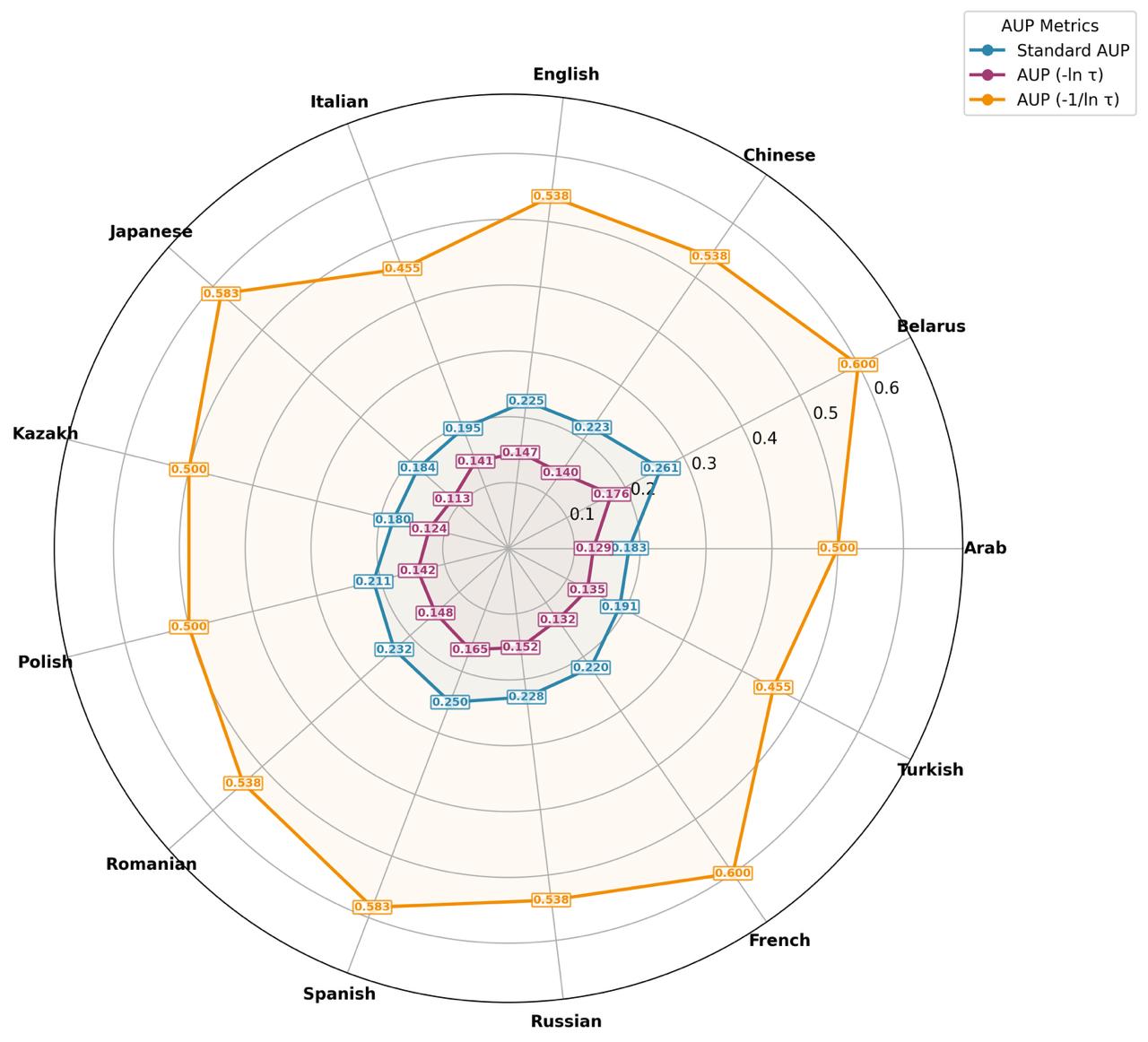}
\end{center}
\caption{Aggregated performance across all languages for the three AUP variants.
}
\label{fig:bench_struct2}
\end{figure}

Together, these metrics provide a more nuanced diagnostic picture than the standard AUP alone. In multilingual evaluation, a model might show similar $\mathrm{AUP}_{\mathrm{std}}$ scores across languages, yet $\mathrm{AUP}_{-\ln\tau}$ could reveal that its advantage in high-resource languages comes from occasional strong wins 
, while $\mathrm{AUP}_{-1/\ln\tau}$ might indicate more stable, incremental improvements in others (Figure~\ref{fig:bench_struct2}).

\section{Extended Analysis}
\label{app:extended_analysis}

\subsection{Cross-Lingual Performance Gap}
\label{app:gap_analysis}

\begin{figure*}[ht!]
\centering
\includegraphics[width=\textwidth]{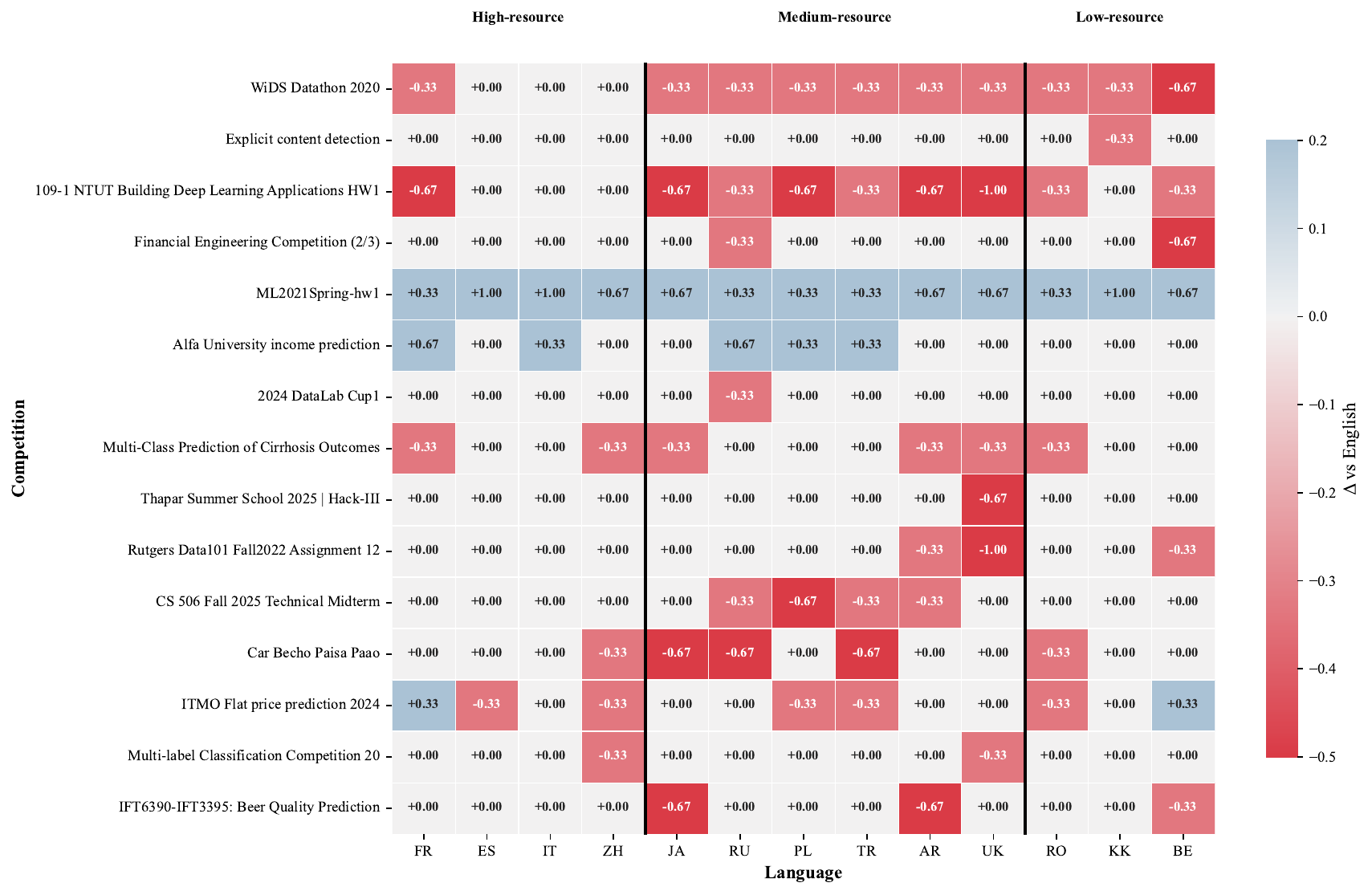}
\caption{Performance gap relative to English across 13 languages for 
GPT-OSS-120b. Negative values indicate worse performance than English.}
\label{fig:gap_english}
\end{figure*}

Figure~\ref{fig:gap_english} shows the per-competition performance gap 
relative to English across 13 languages for GPT-OSS-120b.

\subsection{Qwen2.5-72B-Instruct Preliminary Evaluation}
\label{app:qwen}

We run preliminary experiments with Qwen2.5-72B-Instruct via AIDE 
(1-hour budget) on 3 competitions $\times$ 3 languages (Russian, 
Kazakh, Chinese). Results are shown in 
Table~\ref{tab:qwen_prelim}.

\begin{table}[ht]
\caption{Qwen2.5-72B-Instruct preliminary results (F1 / RMSE / 
Logloss). None indicates a failed submission.}
\centering
\small
\begin{tabular}{lccc}
\toprule
\textbf{Competition (metric)} & \textbf{Russian} & \textbf{Kazakh} & \textbf{Chinese} \\
\midrule
Explicit content (F1)      & 0.941 & 0.971 & 0.967 \\
ML2021Spring-hw1 (RMSE)    & None  & None  & None  \\
ECE460j-fall24 (Logloss)   & None  & 0.495 & None  \\
\bottomrule
\end{tabular}
\label{tab:qwen_prelim}
\end{table}

Qwen fails to produce valid submissions in 5 of 9 pairs, 
with complete failure on the RMSE task. When Qwen succeeds --- on 
the F1 task --- scores are broadly comparable to GPT-OSS-120b 
(0.941--0.971 vs.\ 0.965--0.988). Failure analysis reveals a 
\emph{configuration leakage} pattern: Qwen generates structurally 
correct \texttt{train}/\texttt{predict} functions but additionally 
emits global-scope code with hardcoded paths from its internal AIDE 
test run. When ML$^2$B executes only the required functions in a 
clean environment, stale path references cause runtime failures. 
This is consistent with the broader pattern that pipeline 
instruction-following is the primary bottleneck for weaker models. 
A full evaluation with matched 3-hour budgets across all 14 languages 
will be included in future work.

\subsection{Task-Characteristic Analysis of Cross-Lingual Variance}
\label{app:task_analysis}

For each of the 15 competitions, we code domain, modality, dataset 
size, number of features, description length (tokens), and number of 
domain-specific terms. We compute per-competition cross-lingual 
standard deviation of performance ratios (std of $r_p$ across 14 
languages, aggregated over all four model-framework configurations) 
and correlate these with task characteristics 
(Table~\ref{tab:task_corr}).

\begin{table}[ht]
\caption{Spearman correlations between task characteristics and 
cross-lingual variance ($N$=15). Mean $r_p$ is uncorrelated with 
all characteristics (all $|\rho| < 0.17$, $p > 0.54$).}
\centering
\small
\begin{tabular}{lcc}
\toprule
\textbf{Task characteristic} & \boldmath$\rho$ & \textbf{p} \\
\midrule
Number of features            & +0.50 & 0.055 \\
Domain-specific terms (count) & +0.44 & 0.100 \\
Dataset size (log)            & +0.36 & 0.187 \\
Description length (tokens)   & +0.04 & 0.874 \\
\bottomrule
\end{tabular}
\label{tab:task_corr}
\end{table}

Task characteristics predict \emph{how much} a result varies across 
languages, not how well models perform on average. Low cross-lingual 
variance has two distinct patterns: tasks where all languages succeed 
(e.g., Thapar Summer School, std=0.0005) and tasks where all languages 
fail equally due to underspecification (e.g., Financial Engineering 2/3, 
std=0.034, mean $r_p \approx 0.60$). Stability is therefore not always 
a success signal. These correlations should be interpreted as 
exploratory evidence ($N$=15 limits statistical power) motivating 
targeted follow-up.

\section*{Acknowledgments}
We would like to thank Ivan Yamshchikov for proposing the initial idea of investigating how the
natural language of prompts influences code generation.

Also, we would like to thank the translation assessors for their time and effort: Aldan Yerbalanov,
Artyom Senokosov, Valeria Aitova, Charlene Madida, Cristian Buliga, Deng Yuying, Ekaterina Fedorishcheva,
Eleonore Vissol-Gaudin, Eugene, Giorgio Vedovi, Mikhail Khvorostianyi, Michal Kucharzewski,
O.L., Saraa Ali, Sahra, Zhang Yujing, Jinyu Zhou.

\section*{Declaration on Generative AI}
During the preparation of this work, the authors used Claude Sonnet 4.6 and Grammarly in order to: Grammar and spelling check. After using these tools, the authors reviewed and edited the content as needed and take full responsibility for the publication’s content.

\bibliography{sample-ceur}

@article{Mahowald2024Dissociating,
title={Dissociating language and thought in large language models},
author={Kyle Mahowald and Anna A. Ivanova and I. Blank and Nancy Kanwisher and Joshua B. Tenenbaum and Evelina Fedorenko},
journal={Trends in Cognitive Sciences},
year={2024},
volume={28},
pages={517-540},
doi={10.1016/j.tics.2024.01.011}
}

@article{Gwet2008Computing,title={Computing inter-rater reliability and its variance in the presence of high agreement.},author={K. Gwet},journal={The British journal of mathematical and statistical psychology},year={2008},volume={61 Pt 1},pages={ 29-48 },doi={10.1348/000711006x126600}}

@article{10.1613/jair.1.11854,
author = {Z\"{o}ller, Marc-Andr\'{e} and Huber, Marco F.},
title = {Benchmark and Survey of Automated Machine Learning Frameworks},
year = {2021},
issue_date = {May 2021},
publisher = {AI Access Foundation},
address = {El Segundo, CA, USA},
volume = {70},
issn = {1076-9757},
url = {https://doi.org/10.1613/jair.1.11854},
doi = {10.1613/jair.1.11854},
journal = {J. Artif. Int. Res.},
month = may,
pages = {409–472},
numpages = {64}
}

@inproceedings{
chan2025mlebench,
title={{MLE}-bench: Evaluating Machine Learning Agents on Machine Learning Engineering},
author={Jun Shern Chan and Neil Chowdhury and Oliver Jaffe and James Aung and Dane Sherburn and Evan Mays and Giulio Starace and Kevin Liu and Leon Maksin and Tejal Patwardhan and Aleksander Madry and Lilian Weng},
booktitle={The Thirteenth International Conference on Learning Representations},
year={2025},
url={https://openreview.net/forum?id=6s5uXNWGIh}
}

@inproceedings{bang-etal-2023-multitask,
    title = "A Multitask, Multilingual, Multimodal Evaluation of {C}hat{GPT} on Reasoning, Hallucination, and Interactivity",
    author = "Bang, Yejin  and
      Cahyawijaya, Samuel  and
      Lee, Nayeon  and
      Dai, Wenliang  and
      Su, Dan  and
      Wilie, Bryan  and
      Lovenia, Holy  and
      Ji, Ziwei  and
      Yu, Tiezheng  and
      Chung, Willy  and
      Do, Quyet V.  and
      Xu, Yan  and
      Fung, Pascale",
    editor = "Park, Jong C.  and
      Arase, Yuki  and
      Hu, Baotian  and
      Lu, Wei  and
      Wijaya, Derry  and
      Purwarianti, Ayu  and
      Krisnadhi, Adila Alfa",
    booktitle = "Proceedings of the 13th International Joint Conference on Natural Language Processing and the 3rd Conference of the Asia-Pacific Chapter of the Association for Computational Linguistics (Volume 1: Long Papers)",
    month = nov,
    year = "2023",
    address = "Nusa Dua, Bali",
    publisher = "Association for Computational Linguistics",
    url = "https://aclanthology.org/2023.ijcnlp-main.45/",
    doi = "10.18653/v1/2023.ijcnlp-main.45",
    pages = "675--718"
}

@misc{huang2024mlagentbenchevaluatinglanguageagents,
      title={MLAgentBench: Evaluating Language Agents on Machine Learning Experimentation}, 
      author={Qian Huang and Jian Vora and Percy Liang and Jure Leskovec},
      year={2024},
      eprint={2310.03302},
      archivePrefix={arXiv},
      primaryClass={cs.LG},
      url={https://arxiv.org/abs/2310.03302}, 
}

@misc{padigela2025mldevbenchcomparativeanalysisai,
      title={ML-Dev-Bench: Comparative Analysis of AI Agents on ML development workflows}, 
      author={Harshith Padigela and Chintan Shah and Dinkar Juyal},
      year={2025},
      eprint={2502.00964},
      archivePrefix={arXiv},
      primaryClass={cs.SE},
      url={https://arxiv.org/abs/2502.00964}, 
}

@article{Xuan2025MMLU-ProX,
title={MMLU-ProX: A Multilingual Benchmark for Advanced Large Language Model Evaluation},
author={Weihao Xuan and Rui Yang and Heli Qi and Qingcheng Zeng and Yunze Xiao and Yun Xing and Junjue Wang and Huitao Li and Xin Li and Kunyu Yu and Nan Liu and Qingyu Chen and Douglas Teodoro and Edison Marrese-Taylor and Shijian Lu and Yusuke Iwasawa and Yutaka Matsuo and Irene Li},
journal={ArXiv},
year={2025},
volume={abs/2503.10497},
doi={10.48550/arxiv.2503.10497}
}

@article{qin2025survey,
  title={A survey of multilingual large language models},
  author={Qin, Libo and Chen, Qiguang and Zhou, Yuhang and Chen, Zhi and Li, Yinghui and Liao, Lizi and Li, Min and Che, Wanxiang and Yu, Philip S},
  journal={Patterns},
  volume={6},
  number={1},
  year={2025},
  publisher={Elsevier}
}

@inproceedings{li2025language,
  title={Language ranker: A metric for quantifying llm performance across high and low-resource languages},
  author={Li, Zihao and Shi, Yucheng and Liu, Zirui and Yang, Fan and Payani, Ali and Liu, Ninghao and Du, Mengnan},
  booktitle={Proceedings of the AAAI Conference on Artificial Intelligence},
  volume={39},
  number={27},
  pages={28186--28194},
  year={2025}
}

@inproceedings{huo2025enhancing,
  title={Enhancing non-english capabilities of english-centric large language models through deep supervision fine-tuning},
  author={Huo, Wenshuai and Feng, Xiaocheng and Huang, Yichong and Fu, Chengpeng and Li, Baohang and Ye, Yangfan and Zhang, Zhirui and Tu, Dandan and Tang, Duyu and Lu, Yunfei and others},
  booktitle={Proceedings of the AAAI Conference on Artificial Intelligence},
  volume={39},
  number={23},
  pages={24185--24193},
  year={2025}
}

@article{shaham2024multilingual,
  title={Multilingual instruction tuning with just a pinch of multilinguality},
  author={Shaham, Uri and Herzig, Jonathan and Aharoni, Roee and Szpektor, Idan and Tsarfaty, Reut and Eyal, Matan},
  journal={arXiv preprint arXiv:2401.01854},
  year={2024}
}

@inproceedings{raihan-etal-2025-mhumaneval,
    title = "m{H}uman{E}val - A Multilingual Benchmark to Evaluate Large Language Models for Code Generation",
    author = "Raihan, Nishat  and
      Anastasopoulos, Antonios  and
      Zampieri, Marcos",
    editor = "Chiruzzo, Luis  and
      Ritter, Alan  and
      Wang, Lu",
    booktitle = "Proceedings of the 2025 Conference of the Nations of the Americas Chapter of the Association for Computational Linguistics: Human Language Technologies (Volume 1: Long Papers)",
    month = apr,
    year = "2025",
    address = "Albuquerque, New Mexico",
    publisher = "Association for Computational Linguistics",
    url = "https://aclanthology.org/2025.naacl-long.570/",
    doi = "10.18653/v1/2025.naacl-long.570",
    pages = "11432--11461",
    ISBN = "979-8-89176-189-6"
}

@dataset{ekaterina_trofimova_2024_12700065,
  author       = {Ekaterina Trofimova and
                  Sataev, Emil and
                  Anastasia Drozdova and
                  Polina Guseva and
                  Anna Scherbakova and
                  Andrey Ustyuzhanin and
                  Anastasia Gorodilova and
                  Valeriy Berezovskiy},
  title        = {Code4ML 2.0: a Large-scale Dataset of annotated
                   Machine Learning Code
                  },
  month        = jul,
  year         = 2024,
  publisher    = {Zenodo},
  version      = {2.0},
  doi          = {10.5281/zenodo.12700065},
  url          = {https://doi.org/10.5281/zenodo.12700065},
}

@inproceedings{muennighoff-etal-2023-crosslingual,
    title = "Crosslingual Generalization through Multitask Finetuning",
    author = "Muennighoff, Niklas  and
      Wang, Thomas  and
      Sutawika, Lintang  and
      Roberts, Adam  and
      Biderman, Stella  and
      Le Scao, Teven  and
      Bari, M Saiful  and
      Shen, Sheng  and
      Yong, Zheng Xin  and
      Schoelkopf, Hailey  and
      Tang, Xiangru  and
      Radev, Dragomir  and
      Aji, Alham Fikri  and
      Almubarak, Khalid  and
      Albanie, Samuel  and
      Alyafeai, Zaid  and
      Webson, Albert  and
      Raff, Edward  and
      Raffel, Colin",
    editor = "Rogers, Anna  and
      Boyd-Graber, Jordan  and
      Okazaki, Naoaki",
    booktitle = "Proceedings of the 61st Annual Meeting of the Association for Computational Linguistics (Volume 1: Long Papers)",
    month = jul,
    year = "2023",
    address = "Toronto, Canada",
    publisher = "Association for Computational Linguistics",
    url = "https://aclanthology.org/2023.acl-long.891/",
    doi = "10.18653/v1/2023.acl-long.891",
    pages = "15991--16111"
}

@inproceedings{
ahuja2023mega,
title={{MEGA}: Multilingual Evaluation of Generative {AI}},
author={Kabir Ahuja and Harshita Diddee and Rishav Hada and Millicent Ochieng and Krithika Ramesh and Prachi Jain and Akshay Nambi and Tanuja Ganu and Sameer Segal and Mohamed Ahmed and Kalika Bali and Sunayana Sitaram},
booktitle={The 2023 Conference on Empirical Methods in Natural Language Processing},
year={2023},
url={https://openreview.net/forum?id=jmopGajkFY}
}

@article{wang2022mconala,
  title={MCoNaLa: A benchmark for code generation from multiple natural languages},
  author={Wang, Zhiruo and Cuenca, Grace and Zhou, Shuyan and Xu, Frank F and Neubig, Graham},
  journal={arXiv preprint arXiv:2203.08388},
  year={2022}
}

@misc{wang2025openhandsopenplatformai,
      title={OpenHands: An Open Platform for AI Software Developers as Generalist Agents}, 
      author={Xingyao Wang and Boxuan Li and Yufan Song and Frank F. Xu and Xiangru Tang and Mingchen Zhuge and Jiayi Pan and Yueqi Song and Bowen Li and Jaskirat Singh and Hoang H. Tran and Fuqiang Li and Ren Ma and Mingzhang Zheng and Bill Qian and Yanjun Shao and Niklas Muennighoff and Yizhe Zhang and Binyuan Hui and Junyang Lin and Robert Brennan and Hao Peng and Heng Ji and Graham Neubig},
      year={2025},
      eprint={2407.16741},
      archivePrefix={arXiv},
      primaryClass={cs.SE},
      url={https://arxiv.org/abs/2407.16741}, 
}

@inproceedings{huang-etal-2024-da,
    title = "{DA}-Code: Agent Data Science Code Generation Benchmark for Large Language Models",
    author = "Huang, Yiming  and
      Luo, Jianwen  and
      Yu, Yan  and
      Zhang, Yitong  and
      Lei, Fangyu  and
      Wei, Yifan  and
      He, Shizhu  and
      Huang, Lifu  and
      Liu, Xiao  and
      Zhao, Jun  and
      Liu, Kang",
    editor = "Al-Onaizan, Yaser  and
      Bansal, Mohit  and
      Chen, Yun-Nung",
    booktitle = "Proceedings of the 2024 Conference on Empirical Methods in Natural Language Processing",
    month = nov,
    year = "2024",
    address = "Miami, Florida, USA",
    publisher = "Association for Computational Linguistics",
    url = "https://aclanthology.org/2024.emnlp-main.748/",
    doi = "10.18653/v1/2024.emnlp-main.748",
    pages = "13487--13521"
}

@misc{yang2022dataleak,
      title={Data Leakage in Notebooks: Static Detection and Better Processes}, 
      author={Chenyang Yang and Rachel A Brower-Sinning and Grace A. Lewis and Christian Kästner},
      year={2022},
      eprint={2209.03345},
      archivePrefix={arXiv},
      primaryClass={cs.SE},
      url={https://arxiv.org/abs/2209.03345}, 
}

@article{Apicella_2025,
   title={Don’t push the button! Exploring data leakage risks in machine learning and transfer learning},
   volume={58},
   ISSN={1573-7462},
   url={http://dx.doi.org/10.1007/s10462-025-11326-3},
   DOI={10.1007/s10462-025-11326-3},
   number={11},
   journal={Artificial Intelligence Review},
   publisher={Springer Science and Business Media LLC},
   author={Apicella, Andrea and Isgrò, Francesco and Prevete, Roberto},
   year={2025},
   month=aug }

@article{Sasse_2025,
   title={Overview of leakage scenarios in supervised machine learning},
   volume={12},
   ISSN={2196-1115},
   url={http://dx.doi.org/10.1186/s40537-025-01193-8},
   DOI={10.1186/s40537-025-01193-8},
   number={1},
   journal={Journal of Big Data},
   publisher={Springer Science and Business Media LLC},
   author={Sasse, L. and Nicolaisen-Sobesky, E. and Dukart, J. and Eickhoff, S. B. and Götz, M. and Hamdan, S. and Komeyer, V. and Kulkarni, A. and Lahnakoski, J. M. and Love, B. C. and Raimondo, F. and Patil, Kaustubh R.},
   year={2025},
   month=may }

@inproceedings{lavie-2010-evaluating,
    title = "Evaluating the Output of Machine Translation Systems",
    author = "Lavie, Alon",
    booktitle = "Proceedings of the 9th Conference of the Association for Machine Translation in the Americas: Tutorials",
    month = oct # " 31-" # nov # " 4",
    year = "2010",
    address = "Denver, Colorado, USA",
    publisher = "Association for Machine Translation in the Americas",
    url = "https://aclanthology.org/2010.amta-tutorials.4/"
}

@misc{liu2025mlmasteraiforaiintegrationexploration,
      title={ML-Master: Towards AI-for-AI via Integration of Exploration and Reasoning}, 
      author={Zexi Liu and Yuzhu Cai and Xinyu Zhu and Yujie Zheng and Runkun Chen and Ying Wen and Yanfeng Wang and Weinan E and Siheng Chen},
      year={2025},
      eprint={2506.16499},
      archivePrefix={arXiv},
      primaryClass={cs.AI},
      url={https://arxiv.org/abs/2506.16499}, 
}

@misc{yang2025rdagentllmagentframeworkautonomous,
      title={R\&D-Agent: An LLM-Agent Framework Towards Autonomous Data Science}, 
      author={Xu Yang and Xiao Yang and Shikai Fang and Yifei Zhang and Jian Wang and Bowen Xian and Qizheng Li and Jingyuan Li and Minrui Xu and Yuante Li and Haoran Pan and Yuge Zhang and Weiqing Liu and Yelong Shen and Weizhu Chen and Jiang Bian},
      year={2025},
      eprint={2505.14738},
      archivePrefix={arXiv},
      primaryClass={cs.AI},
      url={https://arxiv.org/abs/2505.14738}, 
}

@article{leakage_in_sci,
title = {Leakage and the reproducibility crisis in machine-learning-based science},
journal = {Patterns},
volume = {4},
number = {9},
pages = {100804},
year = {2023},
issn = {2666-3899},
doi = {https://doi.org/10.1016/j.patter.2023.100804},
url = {https://www.sciencedirect.com/science/article/pii/S2666389923001599},
author = {Sayash Kapoor and Arvind Narayanan}
}

@misc{zhou2025lessleakbench,
      title={LessLeak-Bench: A First Investigation of Data Leakage in LLMs Across 83 Software Engineering Benchmarks}, 
      author={Xin Zhou and Martin Weyssow and Ratnadira Widyasari and Ting Zhang and Junda He and Yunbo Lyu and Jianming Chang and Beiqi Zhang and Dan Huang and David Lo},
      year={2025},
      eprint={2502.06215},
      archivePrefix={arXiv},
      primaryClass={cs.SE},
      url={https://arxiv.org/abs/2502.06215}, 
}

@misc{matton2024leakage,
      title={On Leakage of Code Generation Evaluation Datasets}, 
      author={Alexandre Matton and Tom Sherborne and Dennis Aumiller and Elena Tommasone and Milad Alizadeh and Jingyi He and Raymond Ma and Maxime Voisin and Ellen Gilsenan-McMahon and Matthias Gallé},
      year={2024},
      eprint={2407.07565},
      archivePrefix={arXiv},
      primaryClass={cs.CL},
      url={https://arxiv.org/abs/2407.07565}, 
}

@article{jiang2025aide,
  title={Aide: Ai-driven exploration in the space of code},
  author={Jiang, Zhengyao and Schmidt, Dominik and Srikanth, Dhruv and Xu, Dixing and Kaplan, Ian and Jacenko, Deniss and Wu, Yuxiang},
  journal={arXiv preprint arXiv:2502.13138},
  year={2025}
}

@misc{vakhrushev2022lightautomlautomlsolutionlarge,
      title={LightAutoML: AutoML Solution for a Large Financial Services Ecosystem}, 
      author={Anton Vakhrushev and Alexander Ryzhkov and Maxim Savchenko and Dmitry Simakov and Rinchin Damdinov and Alexander Tuzhilin},
      year={2022},
      eprint={2109.01528},
      archivePrefix={arXiv},
      primaryClass={cs.LG},
      url={https://arxiv.org/abs/2109.01528}, 
}

@article{ouyang2025dscodebench,
  title={DSCodeBench: A Realistic Benchmark for Data Science Code Generation},
  author={Ouyang, Shuyin and Huang, Dong and Guo, Jingwen and Sun, Zeyu and Zhu, Qihao and Zhang, Jie M},
  journal={arXiv preprint arXiv:2505.15621},
  year={2025}
}

@inproceedings{ds100,
author = {Lai, Yuhang and Li, Chengxi and Wang, Yiming and Zhang, Tianyi and Zhong, Ruiqi and Zettlemoyer, Luke and Yih, Wen-tau and Fried, Daniel and Wang, Sida and Yu, Tao},
title = {DS-1000: a natural and reliable benchmark for data science code generation},
year = {2023},
publisher = {JMLR.org},
booktitle = {Proceedings of the 40th International Conference on Machine Learning},
articleno = {756},
numpages = {27},
location = {Honolulu, Hawaii, USA},
series = {ICML'23}
}

@misc{jiao2023chatgptgoodtranslatoryes,
      title={Is ChatGPT A Good Translator? Yes With GPT-4 As The Engine}, 
      author={Wenxiang Jiao and Wenxuan Wang and Jen-tse Huang and Xing Wang and Shuming Shi and Zhaopeng Tu},
      year={2023},
      eprint={2301.08745},
      archivePrefix={arXiv},
      primaryClass={cs.CL},
      url={https://arxiv.org/abs/2301.08745}, 
}

@misc{raunak2023gptsproduceliteraltranslations,
      title={Do GPTs Produce Less Literal Translations?}, 
      author={Vikas Raunak and Arul Menezes and Matt Post and Hany Hassan Awadalla},
      year={2023},
      eprint={2305.16806},
      archivePrefix={arXiv},
      primaryClass={cs.CL},
      url={https://arxiv.org/abs/2305.16806}, 
}

@article{code4ml,
  title={Code4ML: a large-scale dataset of annotated Machine Learning code},
  author={Drozdova, Anastasia and Trofimova, Ekaterina and Guseva, Polina and Scherbakova, Anna and Ustyuzhanin, Andrey},
  journal={PeerJ Computer Science},
  volume={9},
  pages={e1230},
  year={2023},
  publisher={PeerJ Inc.}
}

@article{agtabular,
  title={AutoGluon-Tabular: Robust and Accurate AutoML for Structured Data},
  author={Erickson, Nick and Mueller, Jonas and Shirkov, Alexander and Zhang, Hang and Larroy, Pedro and Li, Mu and Smola, Alexander},
  journal={arXiv preprint arXiv:2003.06505},
  year={2020}
}

@article{tang2024autogluon,
  title={AutoGluon-Multimodal (AutoMM): Supercharging Multimodal AutoML with Foundation Models},
  author={Tang, Zhiqiang and Fang, Haoyang and Zhou, Su and Yang, Taojiannan and Zhong, Zihan and Hu, Tony and Kirchhoff, Katrin and Karypis, George},
  journal={arXiv preprint arXiv:2404.16233},
  year={2024}
}

@article{moumoula2025evaluating,
  title={Evaluating Programming Language Confusion},
  author={Moumoula, Micheline B{\'e}n{\'e}dicte and Kabore, Abdoul Kader and Klein, Jacques and Bissyande, Tegawend{\'e} F},
  journal={arXiv preprint arXiv:2503.13620},
  year={2025}
}

@article{li2024bridging,
  title={Bridging the Language Gap: Enhancing Multilingual Prompt-Based Code Generation in LLMs via Zero-Shot Cross-Lingual Transfer},
  author={Li, Mingda and Mishra, Abhijit and Mujumdar, Utkarsh},
  journal={arXiv preprint arXiv:2408.09701},
  year={2024}
}

@inproceedings{cosma-etal-2024-rocode,
    title = "{R}o{C}ode: A Dataset for Measuring Code Intelligence from Problem Definitions in {R}omanian",
    author = "Cosma, Adrian  and
      Iordache, Ioan-Bogdan  and
      Rosso, Paolo",
    editor = "Calzolari, Nicoletta  and
      Kan, Min-Yen  and
      Hoste, Veronique  and
      Lenci, Alessandro  and
      Sakti, Sakriani  and
      Xue, Nianwen",
    booktitle = "Proceedings of the 2024 Joint International Conference on Computational Linguistics, Language Resources and Evaluation (LREC-COLING 2024)",
    month = may,
    year = "2024",
    address = "Torino, Italia",
    publisher = "ELRA and ICCL",
    url = "https://aclanthology.org/2024.lrec-main.1236/",
    pages = "14173--14185"
}

@misc{dolan2004benchmarking,
      title={Benchmarking Optimization Software with Performance Profiles}, 
      author={Elizabeth D. Dolan and Jorge J. Moré},
      year={2004},
      eprint={cs/0102001},
      archivePrefix={arXiv},
      primaryClass={cs.MS},
      url={https://arxiv.org/abs/cs/0102001}, 
}

@inproceedings{yao2022react,
  title={React: Synergizing reasoning and acting in language models},
  author={Yao, Shunyu and Zhao, Jeffrey and Yu, Dian and Du, Nan and Shafran, Izhak and Narasimhan, Karthik R and Cao, Yuan},
  booktitle={The eleventh international conference on learning representations},
  year={2022}
}


\end{document}